\documentclass[a4paper,12pt,headinclude,footinclude]{book}

\newcommand{\chapterfolders}{
  Introduction,
  LiteratureReview,
  PerturbationAnalysis,
  OptimizationInducedDynamics,
  FourierAnalysis,
  GlobalRobustness,
  FutureWork,
  Conclusions
}%

\usepackage{silence}
\PassOptionsToPackage{hyphens}{url}

\PassOptionsToPackage{
    pdftitle={Principles of Lipschitz continuity in neural networks},
    pdfauthor={R\'ois\'in Luo},
    pdfsubject={Doctoral Thesis},
    pdfkeywords={Lipschitz Continuity, Robustness, Optimization-Induced Stochastic Dynamics, Parameter Perturbation, Deep Learning Theory, Statistics in Deep Learning, Safety and Trustworthiness, Deep Learning},
    pdfcreator={pdflatex}
}{hyperref}

\usepackage[T1]{fontenc}
\usepackage[utf8]{inputenc}

\usepackage[intoc, english]{nomencl}
\usepackage{etoolbox}
\makenomenclature

\usepackage{nameref}

\usepackage[english]{babel}
\usepackage[hyphens]{url}

\usepackage{setspace}
\usepackage{graphicx}

\usepackage{calc}
\usepackage{csquotes}
\usepackage{titling}
\usepackage[inline]{enumitem}
\usepackage[toc,page]{appendix}

\usepackage{multirow}
\usepackage{multicol}

\usepackage[
  algoruled,
  nofillcomment,
  algochapter,
]{algorithm2e}

\usepackage{booktabs}

\usepackage{array}
\usepackage{arydshln}  

\usepackage{caption}
\usepackage{siunitx}
\usepackage{scrextend}

\usepackage[
  eulerchapternumbers,
  eulermath, 
  subfig,
  beramono,
  pdfspacing,
  floatperchapter,
  dottedtoc,
]{classicthesis}

\usepackage{times} 

\usepackage{arsclassica}

\usepackage[  
  left=3cm,
  right=3cm,
  top=3cm,
  bottom=3cm,
  heightrounded,
]{geometry}

\addto\extrasenglish{
  
}

\titlespacing*{\chapter}{0pt}{0pt}{1em}

\usepackage{subfig}
\usepackage{caption}
\usepackage{subcaption}
\usepackage{subfloat}

\usepackage{import}
\usepackage{float}
\usepackage{float}
\usepackage{multirow}

\usepackage{lscape}
\usepackage{listings}
\usepackage{rotating}
\usepackage{tabularx}

\usepackage{ragged2e}

\usepackage{epigraph}

\usepackage{listings}
\usepackage[table,xcdraw]{xcolor}
\usepackage{colortbl}
\usepackage{float}
\usepackage{tabularx}

\usepackage{amsmath}

\usepackage{amssymb}
\usepackage{mathtools}
\usepackage{amsthm} 
\usepackage{mathabx}
\usepackage{bbm}
\usepackage{acronym}
\usepackage{mathrsfs}
\usepackage{physics}
\usepackage{xfrac}
\usepackage{thmtools}

\usepackage{extarrows}

\usepackage{bbold}
\usepackage{cases}
\usepackage{empheq}

\usepackage{lettrine}

\usepackage{calligra} 

\usepackage{tikz}
\usetikzlibrary{positioning,shapes.geometric,arrows.meta}

\usepackage{mdframed}
\usepackage{pgf} 
\usetikzlibrary{calc}

\usepackage{pgfplots}
\pgfplotsset{compat=newest}

\usetikzlibrary{calc,backgrounds,mindmap,decorations.markings,intersections}
\usetikzlibrary{shapes, shadows, arrows, positioning}
\usetikzlibrary{arrows.meta}
\usetikzlibrary{shadows.blur}
\usetikzlibrary{trees, shadows.blur, calc, decorations.pathreplacing}

\usetikzlibrary{decorations.pathmorphing,decorations.pathreplacing,calc,positioning}
\usetikzlibrary{patterns}

\makeatletter
    \newlist{treelist}{itemize}{5}
    \setlist[treelist]{label=\treelist@label}

    \tikzset{treelist line/.style={thick, line cap=round, rounded corners}}
    \def\treelist@label{%
        \begin{tikzpicture}[remember picture, baseline={([yshift=-.6ex] treelist-bullet-\the\enit@depth.center)}]
            \draw [treelist line] (0, 0) -- node (treelist-bullet-\the\enit@depth) {} ++(.5em, 0);
        \end{tikzpicture}%
        \ifnum\enit@depth>1
            \tikz[remember picture, overlay] \draw [treelist line] (treelist-bullet-\the\numexpr\enit@depth-1\relax.center) |- (treelist-bullet-\the\enit@depth.center);%
        \fi
    }
\makeatother

\usepackage{forest}

\usepackage{adjustbox}

\usepackage[most]{tcolorbox}
\newtcolorbox{citebox}{
  breakable,
  enhanced,
  frame hidden,
  interior hidden,
  size=minimal,
  left skip=8pt,
  borderline west={2pt}{-8pt}{lightgray}
}

\newtcolorbox{questionbox}[1]{
    lower separated=true,
    boxrule=0.5pt,
    colback=white,
    colframe=black,
    coltitle=black,
    top=4mm,
    enhanced,
    sharp corners,
    halign=left,
    boxed title style={colframe=white,colback=white,left=3pt,right=3pt},
    attach boxed title to top left={xshift=0.25cm,yshift=-3.5mm},
    title=\textbf{\scriptsize\textsc{#1}}
}

\theoremstyle{plain}

\theoremstyle{definition}

\theoremstyle{remark}

\usepackage[round,compress]{natbib}

\usepackage{etoolbox} 
\usepackage{bibentry} 

\nobibliography*

\newcommand{\fullcite}[1]{%
  \begingroup
    \nocite{#1}
    \renewcommand{\bibname}{} 
    \bibentry{#1} 
  \endgroup
}

\usepackage{xspace}

\newcommand{\ie}{\emph{i.e.}\xspace}
\newcommand{\eg}{\emph{e.g.}\xspace}
\newcommand{\etc}{etc}

\usepackage{titlesec} 
\usepackage{changepage} 
\usepackage{sectsty} 

\usepackage[object=vectorian]{pgfornament}

\def\app#1#2{%
  \mathrel{%
    \setbox0=\hbox{$#1\sim$}%
    \setbox2=\hbox{%
      \rlap{\hbox{$#1\propto$}}%
      \lower1.1\ht0\box0%
    }%
    \raise0.25\ht2\box2%
  }%
}

\usepackage{pgffor} 
\usepackage{graphicx}
\usepackage{rotating} 

\makeatletter
\def\thesisgraphicpaths{
  {./}%
  {./logo/}%
}%

\foreach \folder in \chapterfolders {
  \xdef\thesisgraphicpaths{%
   \thesisgraphicpaths%
   {./main_matter/chapters/\folder/plots/}%
  }%
}

\graphicspath{\thesisgraphicpaths}
\makeatother%

\def\thesisinputpaths{
  {./}%
  {./tikz/}%
}%

\foreach \folder in \chapterfolders {%
  \xdef\thesisinputpaths{%
  \thesisinputpaths%
  {./main_matter/chapters/\folder/}
  {./main_matter/chapters/\folder/tikz/}%
  }
}

\makeatletter 
  \def\input@path{\thesisinputpaths}%
\makeatother%

\usepackage{subfiles} 

\renewcommand{\mkbegdispquote}[2]{\openautoquote}

\SetEnumitemKey{inline}{
  label=(\arabic*)
}

\SetEnumitemKey{research-questions}{
  label=\textbf{(RQ\arabic*)},
  ref=(RQ\arabic*),
  leftmargin=\widthof{\textbf{(RQ0)}}+\labelsep
}

\SetEnumitemKey{secondary-research-questions}{
  label=\textbf{(SRQ\arabic*)},
  ref=(SRQ\arabic*),
  leftmargin=\widthof{\textbf{(SRQ0)}}+\labelsep
}

\SetEnumitemKey{noindent}{
  leftmargin=*
}

\title{Principles of Lipschitz continuity in neural networks}
\author{R\'{o}is\'{i}n Luo}
\date{September, 2025 (v3)}

\begin{document}

\begin{titlepage}

\thispagestyle{empty}

    \begin{center}

    \setlength{\parindent}{0pt}
    \setlength{\parskip}{0pt}

    \begin{tabular}{
        >{\raggedleft\arraybackslash}m{0.2\linewidth} 
        @{\hspace{0.03\linewidth}} 
        >{\raggedright\arraybackslash}m{0.3\linewidth} 
    }
         \includegraphics[trim={3.5cm 12.9cm 3.5cm 3cm},clip,width=1\linewidth]{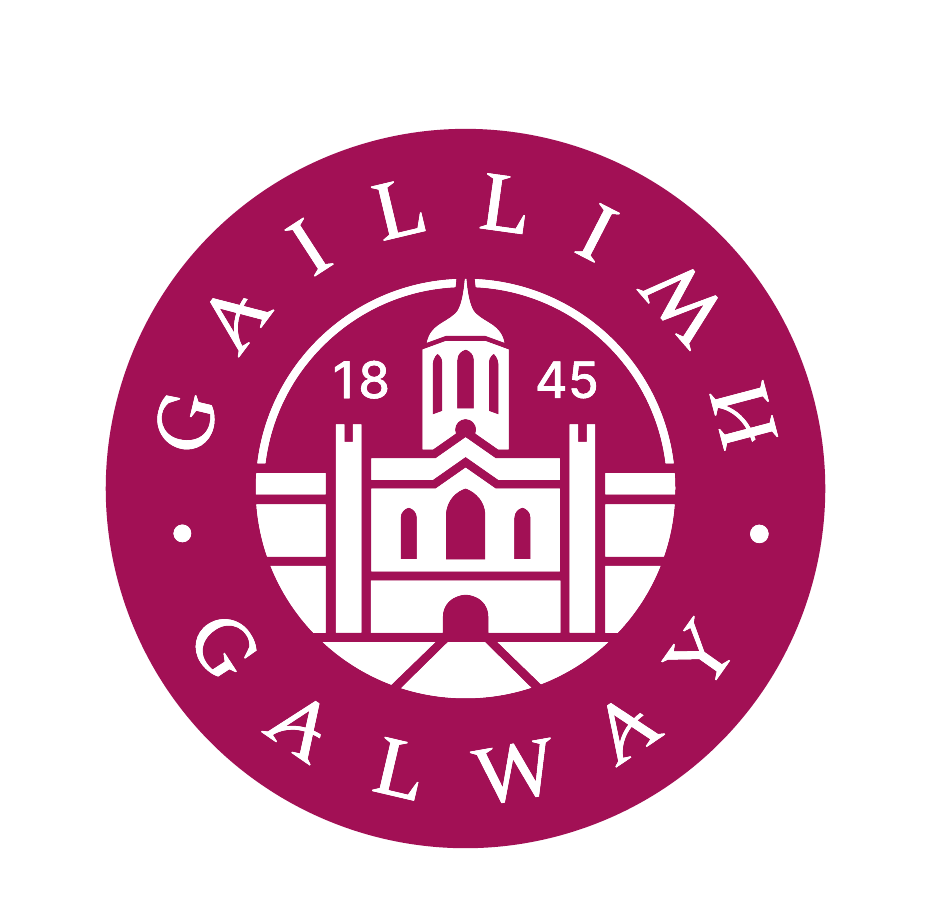}
        &
        \includegraphics[width=1\linewidth]{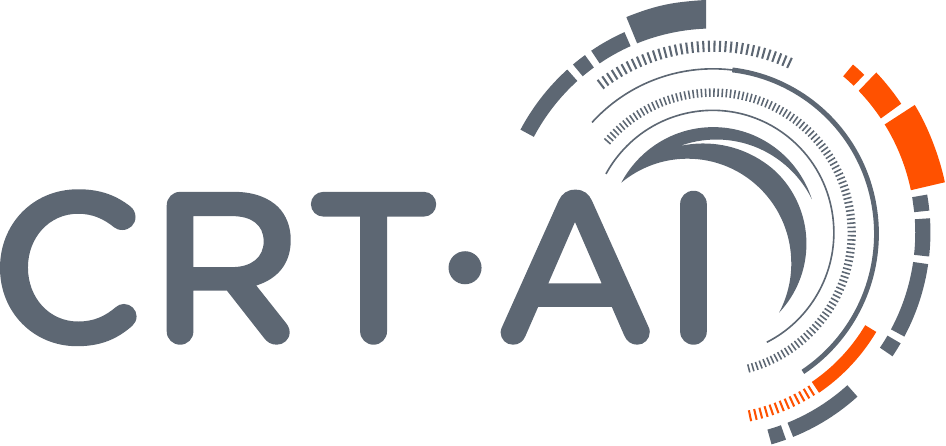}
    \end{tabular}

    \vspace*{\stretch{1}}

    {\huge \textbf{\thetitle}\par}
    
    \vspace{0.5\baselineskip}
    {\large by\par}
    \vspace{0.25\baselineskip}
    {\large \theauthor\par}

    \vspace*{\stretch{1}}

    {\large \textbf{A thesis presented in fulfillment of the requirements for the degree of Doctor of Philosophy}\par}

    \vspace{0.5\baselineskip}
    {\large at \par}

    \vspace{0.5\baselineskip}
    {\large
        Ollscoil na Gaillimhe\\
        University of Galway\\
        Ireland
    \par}

    \vspace{0.5\baselineskip}
    {\large and \par}
    \vspace{0.5\baselineskip}
    
    {\large 
        Taighde \'Eireann -- Research Ireland\\
        Irish National Centre for Research Training in AI (CRT-AI)
    \par}

    \vspace*{\stretch{1}}

    {\large
        \begin{tabular}{rl}
            \textbf{Doctoral Supervisors}: & Dr.\ Colm O'Riordan\\
                                 & Dr.\ James McDermott
        \end{tabular}
    \par}

    \vspace*{\stretch{2}}

    {\large
        \begin{tabular}{m{0.5\textwidth} m{0.5\textwidth}}
            Scoil na R\'iomheola\'iochta \hfill & \hfill \multirow{2}{*}{\thedate} \\
            School of Computer Science 
        \end{tabular}
    \par}

    \end{center}

    
\end{titlepage}
\clearpage 
\thispagestyle{empty} 

      \null\vfill
        
      {
        \large
        \textbf{Cite as:}
        \medskip
        \begin{citebox}
          \textcolor{ForestGreen}{ 
            \theauthor.~\thetitle.~Doctoral Thesis,~Research Ireland -- Centre for Research Training in Artificial Intelligence (CRT-AI),~University of Galway, Ireland,~\thedate.
           }
        \end{citebox}
        \par
      }
      \vfill \vfill
        
\clearpage

\clearpage 
\thispagestyle{empty} 

    \null\vfill

    {
     \vspace{6ex}
    \begin{quotation}
      \begin{center}
        \begin{em}
        {
          \large
          \bfseries
          To my beloved grandparents.
          \par
         }
         \end{em}
         \end{center}
       \end{quotation}
    }

      \vfill \vfill
        
\clearpage

\clearpage

\thispagestyle{empty} 
\null\vfill

\begin{center}

    \setlength{\parindent}{0pt}
    \setlength{\parskip}{0pt}

    \vspace*{\stretch{1}}

    {
      \captionsetup[figure]{labelformat=empty}    
      \begin{figure}[H]
        \centering
        \includegraphics[width=1\linewidth]{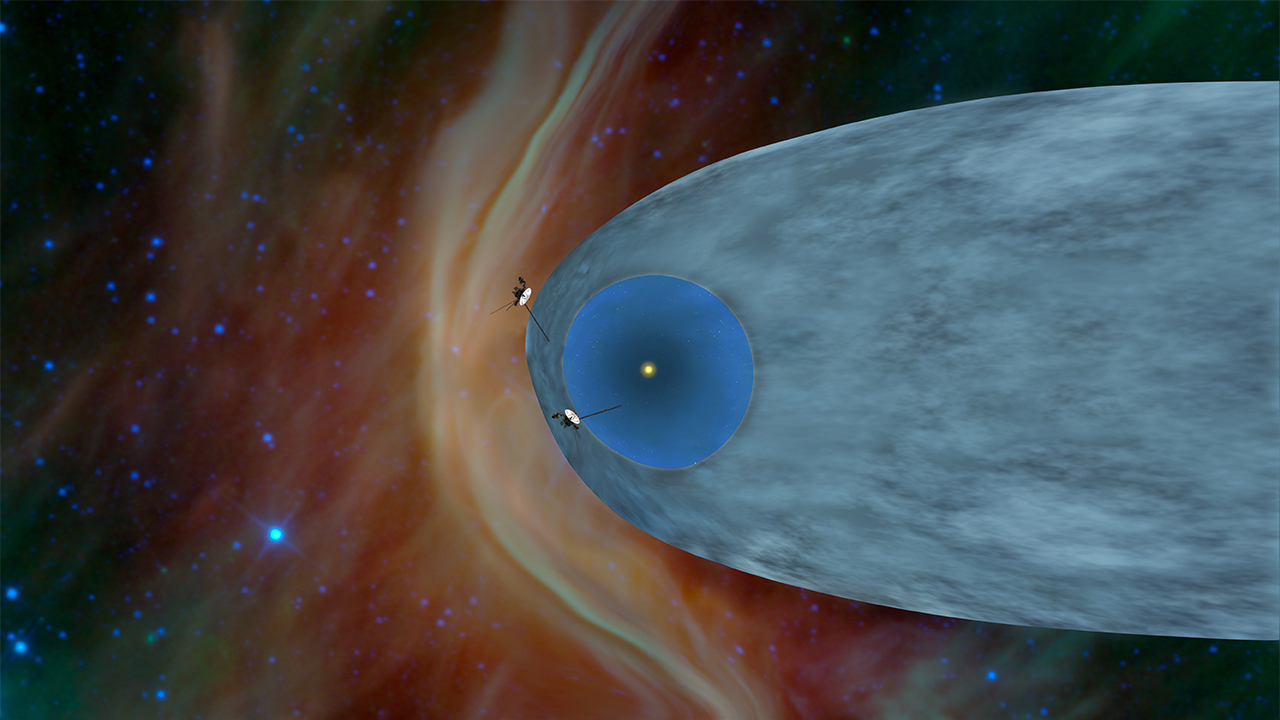}
        
        \caption*{\textbf{Voyagers: Edge of the Bubble} (NASA/JPL-Caltech)\protect\footnotemark{}. Two tiny spacecraft --- the Voyagers --- manifest our spirit of exploration to an extraordinary level, a spirit that defines us as human beings. They have been sailing outward into vast interstellar space, reaching the frontier of our solar system, where the star winds meet the solar winds from our world, forming a vast bubble. Within this bubble lies every story, every life, every dream we have ever known --- yet the Voyagers press onward, carrying our boldness and wonder into the vast, eternal, and spectacular sea between the stars, destined for an eternal journey, a testament to our instinct to go beyond the horizon where no human has gone before.}

      \end{figure} 

    }
    
    \footnotetext{\footnotesize Image use follows the \emph{NASA Images and Media Usage Guidelines}.}
    
    \vspace*{\stretch{1}}

    \begin{quote}
        {
        \centering
         I am only stardust --- learning to think, to chase dreams.
        \par
        }
    \end{quote}
    
    \vspace*{\stretch{1}}
    
\end{center}

\vfill\vfill

\clearpage

\frontmatter
\clearpage

\thispagestyle{empty} 
\null\vfill



\epigraph{``Nunquam praescriptos transibunt sidera fines --- Nothing exceeds the limits of the stars.''}{\textit{Jules Henri Poincar\'e, 1890}}

\vfill\vfill

\clearpage

\chapter*{Research Funding}


This Ph.D.\ training program is a part of the Centre for Research Training in Artificial Intelligence (\textbf{CRT-AI}), funded by \textbf{Taighde \'Eireann -- Research Ireland} under Grant Number 18/CRT/6223.
\chapter*{Abstract}


Deep learning has achieved remarkable success across a wide range of domains, significantly expanding the frontiers of what is achievable in artificial intelligence. Yet, despite these advances, critical challenges remain --- most notably, ensuring \textbf{robustness} to small input perturbations and \textbf{generalization} to out-of-distribution data. These critical challenges underscore the need to understand the underlying fundamental principles that govern robustness and generalization. This understanding is indispensable for establishing deep learning systems that are: \textbf{reliable} --- performing consistently under expected conditions on in-distribution data; \textbf{resilient} --- capable of recovering from unexpected conditions such as noise or adversarial attacks; and \textbf{trustworthy} --- behaving transparently, ethically, in alignment with intended use, and technically robust, particularly in safety-critical applications.

Among the theoretical tools available, \textbf{Lipschitz continuity} plays a pivotal role in governing
the fundamental properties of neural networks related to robustness and generalization. It quantifies the worst-case sensitivity of network's outputs to small input perturbations. While its importance is widely acknowledged, prior research has predominantly focused on empirical regularization approaches based on Lipschitz constraints, leaving the underlying principles less explored. This thesis seeks to advance a principled understanding of the \textbf{principles of Lipschitz continuity in neural networks} within the paradigm of machine learning, examined from \textbf{two complementary perspectives}: an \textbf{internal} perspective --- focusing on the temporal evolution of Lipschitz continuity in neural networks during training (\ie, \textbf{training dynamics}); and an \textbf{external} perspective --- investigating how Lipschitz continuity modulates the behavior of neural networks with respect to features in the input data, particularly its role in governing frequency signal propagation (\ie, \textbf{modulation of frequency signal propagation}).

Guided by these perspectives, the thesis formulates three primary research questions: \textbf{(RQ1) State of Knowledge} --- what is the state of knowledge of Lipschitz continuity in neural networks? \textbf{(RQ2) Training Dynamics} --- how does Lipschitz continuity in neural networks evolve during the training process? \textbf{(RQ3) Modulation of Frequency Signal Propagation} --- how does Lipschitz continuity modulate frequency signal propagation in neural networks? 



\vfill\vfill

\tableofcontents
\listoffigures
\listoftables

\listoftheorems[ignoreall, numwidth=3.5em, onlynamed={theorem},title=List of Theorems]
\listoftheorems[ignoreall, numwidth=3.5em, onlynamed={lemma},title=List of Lemmas]
\listoftheorems[ignoreall, numwidth=3.5em, onlynamed={proposition},title=List of Propositions]
\listoftheorems[ignoreall, numwidth=3.5em, onlynamed={definition},title=List of Definitions]

\chapter*{Declaration}

\vspace{10em}


I, \textit{\theauthor}, declare that this thesis, titled ``\textit{\thetitle}'', is composed by myself, that the work contained herein is my own except where explicitly stated otherwise in the text, and that this work has not been submitted for any other degree or professional qualification.

\vspace{2.5em}
Galway, \thedate

\vspace{5em}
\hfill
\begin{tabular}{c}
	\rule{4cm}{0.5pt} \\
	\theauthor
\end{tabular}

\chapter*{List of publications}

\section*{Articles Compiled in This Thesis}

This thesis compiles and presents the following selected articles --- published or under review as of the date of submission --- into a \underline{\textbf{self-contained}}, \underline{\textbf{modular}}, \underline{\textbf{coherent}}, \underline{\textbf{monograph-style}} presentation format, all centered on the \textbf{principles of Lipschitz continuity in neural networks}:
\begin{enumerate}[label=\Roman*]
    \item \fullcite{luo2025lipschitzsurvey}
    
    \item \fullcite{luo2025spectralvariation}    
    
    \item \fullcite{luo2025lipschitz}

    \item \fullcite{luo2024interpreting}
    
\end{enumerate}

\section*{Articles Not Compiled in This Thesis}

Although the following published conference articles and the manuscript in preparation were produced during the Ph.D. training of the author, they are not included in this thesis due to their topics and the thesis's length constraints:
\begin{enumerate}[resume,label=\Roman*]
    \item \fullcite{luo2024BMVC}


\end{enumerate}

\mainmatter

\nomenclature{$\ell^p,\ell_p$}{A function space of finite dimensions with $p$-norm}

\nomenclature{$L^p,L_p$}{A function space of infinite dimensions with $p$-norm}

\nomenclature{$L^1(\mathbb{R}^d)$}{Absolutely integrable in $\mathbb{R}^d$}

\nomenclature{$L^2(\mathbb{R}^d)$}{Square-integrable in $\mathbb{R}^d$}


\nomenclature{$\displaystyle \int_{\Omega} f\,\mathrm{d}\mu$}{Lebesgue integral of a measurable function $f\colon \Omega \to \mathbb{R}$ with respect to the measure $\mu$ on a measurable set $\Omega$}

\nomenclature{$\displaystyle \int_{\mathbb{R}^d} f(x)\,\mathrm{d}x$}{Lebesgue integral of a measurable (and integrable) function $f\colon \mathbb{R}^d\to\mathbb{R}$ with respect to Lebesgue measure $\mathrm{d}x$ on $\mathbb{R}^d$.}

\nomenclature{$\displaystyle \int f(x)\,dx$}{Indefinite Riemann integral (antiderivative) of $f\colon I\to\mathbb{R}$; any $F$ with $F'(x)=f(x)$ on interval $I$, defined up to an additive constant}

\nomenclature{$\displaystyle \int_a^b f(x)\,dx$}{Riemann integral of a function $f:[a,b]\to\mathbb{R}$ that is Riemann integrable on $[a,b]$}

\nomenclature{$\mathscr{F}[f]$}{Fourier transform of a function $f$}

\nomenclature{$\mathscr{F}^{-1}[F]$}{Inverse Fourier transform of a Fourier transform $F$}

\nomenclature{$\mathbb{R}$}{Set of real numbers}
\nomenclature{$\mathbb{R}^d$}{Euclidean $n$-dimensional space}
\nomenclature{$\mathbb{N}$}{Set of natural numbers}
\nomenclature{$\mathbb{Q}$}{Set of rational numbers}
\nomenclature{$\mathbb{Z}$}{Set of integers}
\nomenclature{$:=$, $\overset{def}{=}$}{Defined as}

\nomenclature{$\equiv$}{Identity or an equivalence relation}

\nomenclature{$\|x\|_p$}{$\ell_p$-norm of $x$}

\nomenclature{$\|T\|_{\mathrm{op}}$}{Operator norm of operator $T$}

\nomenclature{$V \subseteq \mathbb{R}^d$}{A vector space $V$ in $\mathbb{R}^d$}

\nomenclature{$(X,d)$}{A metric space $X$ equipped with a metric $d: X \times X \to \mathbb{R}_+$}


\nomenclature{$f: X \to Y$}{Function from $X$ to $Y$}

\nomenclature{$K$}{Lipschitz constant of a function}
\nomenclature{$K_f$}{Lipschitz constant of a function $f$}
\nomenclature{$K(t)$}{Lipschitz constant of a function over time $t$}
\nomenclature{$K_f(t)$}{Lipschitz constant of a function $f$ over time $t$}

\nomenclature{$\operatorname{Lip}[f]$}{Lipschitz constant of a function $f$}
\nomenclature{$\operatorname{Lip}[f](t)$}{Lipschitz constant of a function $f$ over time $t$}

\nomenclature{$A^\top$}{Transpose of a matrix $A$}
\nomenclature{$A^*$}{Conjugate transpose of a complex matrix $A$}

\nomenclature{$\|T\|_{op}$}{Operator norm of an operator $T$}

\nomenclature{$\|A\|_2$}{Spectral norm of a matrix $A$}

\nomenclature{$\det A, |A|$}{Determinant of a square matrix $A$}

\nomenclature{$\lambda_{\max}(A)$}{Largest eigenvalue of $A$}

\nomenclature{$\mu(A)$}{Lebesgue measure of a measurable set $A\subseteq X$}

\nomenclature{$\operatorname{tr}(A)$}{Trace of a matrix or an operator $A$}
\nomenclature{$\sigma(A)$}{Spectrum of an operator or a matrix $A$}

\nomenclature{$\circledast$}{Convolution operator}
\nomenclature{$f \circ g$}{Composition of two functions $f$ and $g$, defined as $f \circ g(x) := f(g(x))$}

\nomenclature{$A \odot B$}{Hadamard product of $A$ and $B$}

\nomenclature{$A \otimes B$}{Tensor product of $A$ and $B$}

\nomenclature{$\sup$}{Supremum}
\nomenclature{$\inf$}{Infimum}
\nomenclature{$\max$}{Maximum}
\nomenclature{$\min$}{Minimum}
\nomenclature{$s.t.$}{Subject to}


\nomenclature{$\overset{d}{\to}$}{Convergence in distribution (weak convergence) of random variables}

\nomenclature{$\mathbb{E}[\cdot]$}{Expectation operator}

\nomenclature{$\exp(x)$}{Exponential function}
\nomenclature{$\log(x)$}{Natural logarithm}

\nomenclature{$\langle u, v\rangle$}{Inner product of $u$ and $v$}
\nomenclature{$\nabla_x f(x)$}{Gradient of a scalar function $f(x)$ with respect to $x$}

\nomenclature{$D^{\alpha}f$}{$\alpha$-times Fr\'echet derivative of a matrix to scalar function $f: \mathbb{R}^{m \times n} \to \mathbb{R}$}
\nomenclature{$D^{\alpha}F$}{$\alpha$-times Fr\'echet derivative of a matrix to matrix function $F: \mathbb{R}^{m \times n} \to \mathbb{R}^{k \times l}$}


\nomenclature{$\mathrm{I}_n$}{$n\times n$ identity matrix}
\nomenclature{$\mathbb{1}_n$}{$n$-dimensional vector of all ones}

\nomenclature{$\mathcal{L}(X,Y)$}{Space of bounded linear operators from $X$ to $Y$}

\nomenclature{$\mathcal{L}(X)$}{Space of bounded linear operators from $X$ to $X$}

\nomenclature{$B_X(x,r)$}{Open ball in $X$ of radius $r$ centered at $x$}
\nomenclature{$C^k(X)$}{Space of $k$-times continuously differentiable functions on $X$}
\nomenclature{$C^{0,1}(X)$}{Space of Lipschitz functions on $X$}



\nomenclature{$\mathop{\mathrm{arg\,min}}_{x \in S} f(x)$}{Values of $x \in S$ where $f(x)$ attains its minimum}

\nomenclature{$\mathop{\mathrm{arg\,max}}_{x \in S} f(x)$}{Values of $x \in S$ where $f(x)$ attains its maximum}
\printnomenclature


\foreach \folder in \chapterfolders {%

\chapter{Introduction}

\label{chapter:introduction}



\lettrine[lines=3]{D}{eep} learning \citep{goodfellow2016deep} has achieved remarkable success across a diversity of domains, including computer vision \citep{krizhevsky2012imagenet,he2016deep,dosovitskiy2020image}, natural language processing \citep{hochreiter1997long,vaswani2017attention,radford2019language,devlin2019bert,brown2020language,openai2023gpt4,deepseekai2025deepseekr1}, and graph representation learning \citep{perozzi2014deepwalk,velivckovic2018graph,xu2018powerful}. These successes have largely expanded the frontiers of what is achievable in artificial intelligence, enabling powerful capabilities across a wide range of tasks, as well as modeling paradigms such as generative modeling \citep{hinton2006reducing,goodfellow2014generative,dinh2017density,ho2020denoising,song2020score,lipman2023flow}. 

Despite these remarkable advances, critical challenges persist, notably in ensuring \textbf{robustness} to small input perturbations and \textbf{generalization} to out-of-distribution data \citep{szegedy2014intriguing,goodfellow2014explaining,hendrycks2019benchmarking,ilyas2019adversarial,recht2019imagenet,luo2024interpreting}. These critical challenges highlight the need to understand the underlying fundamental principles that govern \textbf{robustness} and \textbf{generalization}, and to establish deep learning systems that are \textbf{reliable} --- performing consistently under expected conditions on in-distribution data \citep{nist2023ai}, \textbf{resilient} --- capable of recovering from unexpected conditions such as noise or adversarial attacks \citep{nist2023ai}, and \textbf{trustworthy} --- behaving transparently, ethically, in alignment with intended use, and technically robust \citep{eu-trustworthy-ai-guidelines,eu-ai-act-2024} --- in safety-critical applications.

\begin{figure}[t!]
    \centering
    
    \resizebox{1\linewidth}{!}{
      \begin{tikzpicture}[
  node distance=2cm,
  databox/.style={
    cylinder,
    shape border rotate=90,
    draw,
    minimum width=2cm,
    minimum height=2.5cm
  },
  funcbox/.style={
    rectangle,
    draw,
    thick,
    minimum width=8cm,
    minimum height=4.5cm
  },
  ellipseblock/.style={
    ellipse,
    draw,
    thick,
    minimum width=2.5cm,
    minimum height=2.5cm
  },
  circleblock/.style={
    circle,
    draw,
    thick,
    minimum size=2cm
  },
  arrow/.style={
        -{Classical TikZ Rightarrow[length=3mm]},
        thick
    }
]

  %
  \makeatletter
  \newcommand{\gear}[5]{%
    \foreach \i in {1,...,#1}{%
      [rotate=(\i-1)*360/#1]
        (0:#2) arc (0:#4:#2)
        -- (#4+#5:#3) arc (#4+#5:360/#1-#5:#3)
        -- (360/#1:#2)
    }%
  }
  \makeatother
  %

  \node[databox]                   (data)      {Data: $x \in \mathcal{X}$};
  
  \node[below=0.3cm of data, text width=8cm, align=center,text=Emerald] (lipschitztext) {\textbf{Lipschitz continuity modulates data feature propagation in $f$.}};
  
  \node[funcbox, right=of data] (function) {}; 
  \node at ([xshift=0cm, yshift=1.4cm]function.center) {Learned function: $f(x;\boldsymbol{\theta}): \mathcal{X} \times \Theta \to \tilde{\mathcal{Y}}$};

    \node[
  anchor=north,
  text width=10cm,
  align=center,text=Emerald
    ] at ([yshift=-0.3cm]function.south) 
    {\textbf{Optimization induces the dynamics of Lipschitz continuity in $f$.}};

  \node[ellipseblock, above=of function]     (prior)     {Supervision signal: $y \in \mathcal{Y}$};
  \node[circleblock,  right=of function] (criterion) {Criterion: $\ell_f(f(x;\boldsymbol{\theta}),y)$};

  \draw[arrow] (data)     -- (function);
  \draw[arrow] (function) -- (criterion);
  \draw[arrow] ([yshift=-1cm]data.north)     |- (prior);
  \draw[arrow] (prior)    -| (criterion);

  \begin{scope}[shift={(function.center)}]
    \def\nteeth{18}
    \def\rin{0.8}     
    \def\rout{1}      
    \def\toothang{10} 
    \def\gapang{2}    
    \def\halfPitch{180/\nteeth}

    \coordinate (G1) at (-2*\rout,0);
    \coordinate (G2) at (   0,   -1cm);
    \coordinate (G3) at ( 2*\rout,0);

    \begin{scope}[fill=gray!50, thick]
      \draw[shift={(G1)}] \gear{\nteeth}{\rin}{\rout}{\toothang}{\gapang};
      \draw[shift={(G2)},rotate=\halfPitch]
        \gear{\nteeth}{\rin}{\rout}{\toothang}{\gapang};
      \draw[shift={(G3)}] \gear{\nteeth}{\rin}{\rout}{\toothang}{\gapang};
    \end{scope}

    \node at (G1) {$\boldsymbol{\theta}^{(1)}$};
    \node at (G2) {$\boldsymbol{\theta}^{(2)}$};
    \node at (G3) {$\boldsymbol{\theta}^{(3)}$};
  \end{scope}

\end{tikzpicture}
    }

    \caption[Principles of Lipschitz Continuity Manifest in Paradigm of Machine Learning]{\textbf{Principles of Lipschitz Continuity Manifest in Paradigm of Machine Learning}. In the paradigm of machine learning, we consider \textit{data} $x \in \mathcal{X}$, \textit{supervision signal} $y \in \mathcal{Y}$, a \textit{learned function} $f: \mathcal{X} \times \boldsymbol{\Theta} \to \Tilde{\mathcal{Y}}$ --- where $\Tilde{\mathcal{Y}}$ is output space --- parameterized by $\theta^{(\ell)} \in \mathbb{R}^{d_\ell}$ and $\boldsymbol{\Theta}:=\mathbb{R}^{d_1 \times d_2 \times \cdots \times d_L}$, and a \textit{criterion} $\ell_f: (f(x, \boldsymbol{\theta}), \mathcal{Y}) \to \mathbb{R}$. The Lipschitz constant $K_f$ of $f$ in this paradigm is determined by factors such as the hypothesis class for $f$, learning algorithm, parameters $\boldsymbol{\theta}$, training data $(\mathcal{X}, \mathcal{Y})$, \etc. It also modulates the behavior of $f$ with respect to the input features. This thesis investigates the \textbf{principles of Lipschitz continuity} in this paradigm from a \textbf{complementary dichotomy of perspectives}: an \textbf{internal} perspective --- focusing on the temporal evolution of Lipschitz continuity in neural networks during training; and an \textbf{external} perspective --- examining how Lipschitz continuity modulates the behavior of $f$ with respect to the features in input data, particularly its role in governing frequency signal propagation in neural networks.}

    \label{fig:paradigm_of_machine_learning}

\end{figure}

The paradigm of machine learning \citep{goodfellow2016deep}, illustrated in Figure~\ref{fig:paradigm_of_machine_learning}, typically comprises the \emph{data} $x \in \mathcal{X} \subseteq \mathbb{R}^m$, \emph{supervision signal} $y \in \mathcal{Y} $, a \emph{learned function} $f(x; \boldsymbol{\theta})$ with output space $\Tilde{\mathcal{Y}} \subseteq \mathbb{R}^n$:
\begin{align}
    f: \mathcal{X} \times \boldsymbol{\Theta} \to \Tilde{\mathcal{Y}}
    ,
\end{align}
parameterized by:
\begin{align}
\boldsymbol{\theta}:=\{\boldsymbol{\theta}^{(1)}, \boldsymbol{\theta}^{(2)},\cdots \} \in \boldsymbol{\Theta}
\subseteq \mathbb{R}^p
,
\end{align}
and a \emph{criterion}:
\begin{align}
    \ell_f: (f(x; \boldsymbol{\theta}), y) \to \mathbb{R}.
\end{align}
Note that the \emph{supervision signals} differ across learning paradigms, for example:
\begin{enumerate}[label=(\roman*)]
    \item \textbf{Supervised Learning.} The \emph{supervision signals} are explicitly provided as \emph{labels} by the dataset;

    \item \textbf{Self-Supervised Learning.} The \emph{supervision signals} are derived from inherent structure or priors in the data --- for example, in contrastive learning, an augmented view of an image can serve as the label for the original image \citep{chen2020simple};

    \item \textbf{Reinforcement Learning.} The \emph{supervision signals} are manifested as \emph{reward signals} from the interactions with the environment, which are often noisy, sparse, and delayed \citep[Preface \& \S~1]{sutton1998reinforcement}.

\end{enumerate}

Among the theoretical tools available, \textbf{Lipschitz continuity} \citep[Theorem 9.19, \S~9]{rudin1976principles} plays a pivotal role in governing the fundamental properties of the function $f$ in this paradigm related to robustness and generalization. The \emph{Lipschitz constant} $K_f$ of $f$, with respect to the $\ell_2$-norm, defined as:
\begin{align}
    K_f := \sup_{\forall~u \neq v} \frac{\|f(u) - f(v)\|_2}{\|u - v\|_2},
\end{align}
measures the worst-case sensitivity of the network to small input perturbations. Equivalently, it bounds the maximum rate of change of $f$ with respect to its inputs in the $\ell_2$-norm \citep[Theorem 9.19, \S~9]{rudin1976principles}:
\begin{align}
K_f \geq \sup_x \|\nabla_x f(x) \|_2
    ,
\end{align}
where $\nabla_x (\cdot)$ denotes a differential operator acting on $(\cdot)$ with respect to $x$. A lower Lipschitz constant $K_f$ ensures a smoother mapping $f$, enhancing the network's ability to generalize to unseen data. Moreover, Lipschitz continuity forms a theoretical foundation for \textbf{certifiable robustness} --- where safety guarantees under bounded perturbations can be formally proven --- in safety-critical applications. 

While Lipschitz continuity is crucial in governing the fundamental properties of neural networks related to robustness and generalization, prior research has largely concentrated on empirical aspects, such as regularization techniques \citep{miyato2018spectral,arjovsky2017wasserstein,cisse2017parseval,anil2019sorting}, leaving the underlying principles less explored. This thesis is motivated by the absence of theoretical underpinnings of the principles of Lipschitz continuity in neural networks --- a foundation that is critical for establishing reliable, resilient, and trustworthy deep learning systems. To address this gap, it seeks to advance a principled understanding of the \textbf{principles of Lipschitz continuity in neural networks} within the paradigm of machine learning, examined from \textbf{a complementary dichotomy of perspectives}: 
\begin{enumerate}[label={}]
    \item an \textbf{internal} perspective --- focusing on the temporal evolution of Lipschitz continuity in neural networks during training (\ie, \textbf{Training Dynamics});

    \item an \textbf{external} perspective --- investigating how Lipschitz continuity modulates the behavior of neural networks with respect to features in the input data, particularly its role in governing frequency signal propagation (\ie, \textbf{Modulation of Frequency Signal Propagation}).
\end{enumerate}
Together, these complementary perspectives form the basis of the primary research questions presented in Section~\ref{sec:research_questions} and underscore the principles under investigation within this thesis.

\section{Research questions}
\label{sec:research_questions}


This thesis is centered around an overarching inquiry from a complementary dichotomy of \textbf{internal} and \textbf{external} perspectives:
\begin{quote}
\textit{What are the \textbf{principles of Lipschitz continuity in neural networks}, from its temporal evolution during training (an \textbf{internal perspective}) to its modulation of frequency signal propagation (an \textbf{external perspective})?}
\end{quote}

To support this central inquiry, it investigates three primary research questions (RQs) with a structured approach, each gives rise to a set of secondary research questions (SRQs), which are explored in one or two dedicated chapters.

\begin{questionbox}{(RQ1) State of Knowledge}
\begin{treelist}
    \item (RQ1) \textbf{State of Knowledge} --- what is the state of knowledge of Lipschitz continuity in deep learning? 
    
    \begin{treelist}
        \item (SRQ1) \textbf{Theoretical Foundations} --- what are the theoretical foundations of Lipschitz continuity in deep learning? 
        
        \item (SRQ2) \textbf{Regularization Mechanism} --- how does the constraint of Lipschitz continuity serve as a regularization mechanism for improving robustness and generalization capabilities in deep learning?
        
    \end{treelist}
\end{treelist}
\end{questionbox}

\begin{questionbox}{(RQ2) Training Dynamics --- An Internal Perspective}
\begin{treelist}
    \item (RQ2) \textbf{Training Dynamics} --- how does Lipschitz continuity of neural networks evolve over time during training?
    \begin{treelist}
        \item (SRQ3) \textbf{Parameter Space Perturbation} --- how do the spectral properties (\eg~singular values) of parameter matrices change in training? During training, the optimization process updates parameter matrices, thereby altering the spectral structures accordingly. 

        \item (SRQ4) \textbf{Stochastic Dynamic Analysis of Lipschitz Continuity} --- how does Lipschitz continuity evolve over time during training in neural networks? 
        
    \end{treelist}
\end{treelist}
\end{questionbox}

\begin{questionbox}{(RQ3) Modulation of Frequency Signal Propagation --- An External Perspective}
\begin{treelist}
    \item (RQ3) \textbf{Modulation of Frequency Signal Propagation} --- how does Lipschitz continuity modulate the frequency signal propagation in neural networks, and how does it help us for understanding the global robustness of image models?
    \begin{treelist}
        \item (SRQ5) \textbf{Fourier Analysis of Lipschitz Continuity} --- how does Lipschitz continuity modulate frequency signal propagation in image models?
        
        \item (SRQ6) \textbf{Interpreting Global Robustness} --- how can global robustness be explained through the lens of spectral bias in Lipschitz continuity? 
    \end{treelist}
\end{treelist}
\end{questionbox}

\section{Thesis Presentation Format}

This thesis compiles and presents a selection of research articles --- published or under review as of the date of submission --- into a \underline{\textbf{self-contained}}, \underline{\textbf{modular}}, \underline{\textbf{coherent}}, \underline{\textbf{monograph-style}} presentation format, all centered on the principles of Lipschitz continuity in neural networks. The research was conducted cumulatively during the Ph.D. training. Portions of the chapter content, which address the research questions, are reused either verbatim or with minor modifications from the selected articles, in accordance with institutional guidelines and the policies of the relevant publishers. Each core chapter is presented in the style of a standalone journal article, including an \textbf{abstract}, \textbf{introduction}, and other standard components in journal articles. For consistency with the original publications, the term \textbf{paper} is used throughout to refer to the current chapter, and the literature review may include results from our research, either published or available as preprints. Permissions for reuses have been obtained from all co-authors in all articles.

\section{Content Organization}

\begin{figure}[t!]
    \centering
    
    \resizebox{1\linewidth}{!}{





\begin{tikzpicture}[node distance=10mm and 15mm, >=Stealth,
  rq1/.style={
        rectangle, draw, rounded corners, 
        fill=green!20, 
        align=center, text width=5cm, minimum height=2cm},
  rq2/.style={
        rectangle, draw, rounded corners, 
        fill=orange!20, 
        align=center, text width=5cm, minimum height=2cm},
  rq3/.style={
        rectangle, draw, rounded corners, 
        fill=cyan!20, 
        align=center, text width=5cm, minimum height=2cm},
  legendbox/.style={
        rectangle, draw, rounded corners, minimum width=12mm, minimum height=12mm, inner sep=0pt},
  legend label/.style={
        anchor=west, 
        }
]

  \begin{scope}[local bounding box=diagram]
  
    \node[rq1] (LiteratureReview) {\hyperref[chapter:literature_review]{Chapter~\ref{chapter:literature_review}: Literature Review}};

    \node[rq1, ellipse, draw, align=center, text width=3.5cm, minimum height=2cm, right=of LiteratureReview, yshift=15mm] (TheoreticalFoundations) {Theoretical Foundations};

    \node[rq1, ellipse, draw, align=center, text width=3.5cm, minimum height=2cm, right=of LiteratureReview, yshift=-15mm] (Regularizations) {Regularizations};

    \node[rq2, above=of TheoreticalFoundations] (PerturbationAnalysis) {\hyperref[chapter:perturbation_analysis]{Chapter~\ref{chapter:perturbation_analysis}: Operator-Theoretic Perturbation Analysis}};

    \node[rq2, ellipse, draw, align=center, text width=3.5cm, minimum height=2cm, left=of PerturbationAnalysis, yshift=15mm] (PerturbationTheory) {Perturbation Theory};

    \node[rq2, ellipse, draw, align=center, text width=3.5cm, minimum height=2cm, left=of PerturbationAnalysis, yshift=-15mm] (OperatorTheory) {Operator Theory};

    \node[rq2, right=of TheoreticalFoundations] (InducedDynamics) {\hyperref[chapter:optimization_induced_dynamics]{Chapter~\ref{chapter:optimization_induced_dynamics}: Stochastic Dynamic Analysis of Lipschitz Continuity}};

    \node[rq2, ellipse, draw, align=center, text width=3.5cm, minimum height=2cm, above=of InducedDynamics] (SDEFoundations) {Matrix-Valued SDE};

    \node[rq2, ellipse, draw, align=center, text width=3.5cm, minimum height=2cm, below=of InducedDynamics] (SGDOptimization) {SGD Optimization};

    \node[rq2, ellipse, draw, align=center, text width=3.5cm, minimum height=2cm, right=of InducedDynamics] (TheoreticalImplications) {Theoretical Implications in Training Dynamics};

    \node[rq2, ellipse, draw, align=center, text width=3.5cm, minimum height=2cm, right=of TheoreticalImplications] (NoisyGradientRegularizations) {Noisy Gradient Regularizations};

    \node[rq2, ellipse, draw, align=center, text width=3.5cm, minimum height=2cm, above=of NoisyGradientRegularizations] (NearConvergenceBehaviours) {Near Convergence Behaviours};

    \node[rq2, ellipse, draw, align=center, text width=3.5cm, minimum height=2cm, above=of NearConvergenceBehaviours] (ParameterInitialization) {Robustness due to Parameter Initialization};

    \node[rq2, ellipse, draw, align=center, text width=3.5cm, minimum height=2cm, below=of NoisyGradientRegularizations] (BatchSize) {Batch Size};

    \node[rq2, ellipse, draw, align=center, text width=3.5cm, minimum height=2cm, below=of BatchSize] (SamplingTrajectory) {Mini-Batch Sampling Trajectory};
    
    \node[rq3, below=of Regularizations] (LocalLipschitz) {\hyperref[chapter:fourier_analysis_of_lipschitz_continuity]{Chapter~\ref{chapter:fourier_analysis_of_lipschitz_continuity}: Fourier Analysis of Lipschitz Continuity}};

    \node[rq3, right=of LocalLipschitz] (GlobalInterpretation) {\hyperref[chapter:interpreting_global_robustness]{Chapter~\ref{chapter:interpreting_global_robustness}: Interpreting Global Robustness}};

    \node[rq3, ellipse, draw, align=center, text width=3.5cm, minimum height=2cm, below=of GlobalInterpretation, xshift=-30mm] (GameTheory) {Shapley Value Theory};

    \node[rq3, ellipse, draw, align=center, text width=3.5cm, minimum height=2cm, below=of GlobalInterpretation, xshift=30mm] (InformationTheory) {Information Theory};

    \draw[->] (LiteratureReview) -- (TheoreticalFoundations);
    \draw[->] (LiteratureReview) -- (Regularizations);
    \draw[->] (OperatorTheory) -- (PerturbationAnalysis);
    \draw[->] (PerturbationTheory) -- (PerturbationAnalysis);
    \draw[->] (TheoreticalFoundations) -- (InducedDynamics);
    \draw[->] (SGDOptimization) -- (InducedDynamics);
    \draw[->] (SDEFoundations) -- (InducedDynamics);
    \draw[->] (PerturbationAnalysis) -- (InducedDynamics);
    \draw[->] (InducedDynamics) -- (TheoreticalImplications);
    
    \draw[->] (TheoreticalImplications) -- (NoisyGradientRegularizations);
    \draw[->] (TheoreticalImplications) -- (NearConvergenceBehaviours);
    \draw[->] (TheoreticalImplications) -- (ParameterInitialization);
    \draw[->] (TheoreticalImplications) -- (BatchSize);
    \draw[->] (TheoreticalImplications) -- (SamplingTrajectory);
    
    \draw[->] (Regularizations) -- (LocalLipschitz);
    \draw[->] (LocalLipschitz) -- (GlobalInterpretation);
    \draw[->] (GameTheory) -- (GlobalInterpretation);
    \draw[->] (InformationTheory) -- (GlobalInterpretation);

    \path (current bounding box.north east) coordinate (bbne);
    \path (current bounding box.south west) coordinate (bbsw);
    \path[use as bounding box] (bbsw) rectangle (bbne);
    
  \end{scope}

  \begin{scope}[local bounding box=legend, 
                above=of diagram,
                anchor=south,
                xshift=50mm,
                yshift=90mm
                ]
    \node[legendbox, fill=green!20] (legC) {};
    \node[legend label, right=1mm of legC] (labC) {RQ1: \textbf{State of Knowledge}};
    
    \node[legendbox, fill=orange!20, right=5cm of legC] (legO) {};
    \node[legend label, right=1mm of legO] (labO) {RQ2: \textbf{Training Dynamics}};
    
    \node[legendbox, fill=cyan!20, right=5cm of legO] (legG) {};
    \node[legend label, right=1mm of legG] (labG) {RQ3: \textbf{Modulation of Frequency Signal Propagation}};
  \end{scope}
  
\end{tikzpicture}

    }
    
    \caption[Thesis Thematic Structure]{\textbf{Thesis Thematic Structure}. A high-level conceptual diagram outlining the thematic connections between chapters. This thesis investigates the principles of Lipschitz continuity in neural networks through a structured approach guided by three primary research questions. Each primary question gives rise to two secondary research questions and is addressed in one or two dedicated chapters.}

    \label{fig:thesis_thematic_structure}
\end{figure}


The research questions collectively establish a cohesive thematic investigation into the principles of Lipschitz continuity in neural networks. Figure~\ref{fig:thesis_thematic_structure} illustrates how each question aligns with the core chapters and jointly guides the development of the overarching theoretical framework. The structure of the thesis is outlined below.

\paragraph{(RQ1) State of Knowledge.} What is the state of knowledge of Lipschitz continuity in deep learning? This research question is addressed by surveying the landscape of Lipschitz continuity in deep learning, covering its theoretical foundations, estimation methods, regularization approaches, and role in certifiable robustness. This research question is addressed by surveying the landscape of Lipschitz continuity in deep learning, covering its theoretical foundations, estimation methods, regularization approaches, and role in certifiable robustness. 
\begin{enumerate}[secondary-research-questions,series=SecondaryResearchQuestions,left=2em]
        \item \textbf{Theoretical Foundations} (\hyperref[chapter:literature_review]{Chapter~\ref{chapter:literature_review}: Literature Review}) --- what are the theoretical foundations of Lipschitz continuity in deep learning? This is answered in \hyperref[chapter:literature_review]{Chapter~\ref{chapter:literature_review}: Literature Review}, which is based on an article:
        \begin{enumerate}[label=(\roman*),series=CompiledPublications]
            \item \textcolor{Emerald}{\fullcite{luo2025lipschitzsurvey}}.
        \end{enumerate}
    
        \item \textbf{Regularization Mechanism} (\hyperref[chapter:literature_review]{Chapter~\ref{chapter:literature_review}: Literature Review}) --- how does Lipschitz continuity serve as a regularization mechanism for improving robustness and generalization capabilities in deep learning? This is answered in \hyperref[chapter:literature_review]{Chapter~\ref{chapter:literature_review}: Literature Review}, which is based on an article:
        \begin{enumerate}[label=(\roman*),series=CompiledPublications]
            \item \textcolor{Emerald}{\fullcite{luo2025lipschitzsurvey}}.
        \end{enumerate}
    \end{enumerate}

\paragraph{(RQ2) Training Dynamics.} How does Lipschitz continuity evolve over time during the training in neural networks? To the best of our knowledge, as of the date of submission, this constitutes the first theoretical framework to characterize the temporal evolution of Lipschitz continuity in neural networks leveraging perturbation theory \citep{kato1995perturbation,luo2025spectralvariation} and a system of stochastic differential equations (SDEs). 
\begin{enumerate}[secondary-research-questions,resume=SecondaryResearchQuestions,left=2em]
        \item \textbf{Parameter Space Perturbation} (\hyperref[chapter:perturbation_analysis]{Chapter~\ref{chapter:perturbation_analysis}: Operator-Theoretic Perturbation Analysis}) --- how do the spectral properties (\ie~singular values) of parameter matrices change in training? During training, the optimization process updates parameter matrices, thereby altering the spectral structures accordingly. This is addressed in \hyperref[chapter:perturbation_analysis]{Chapter~\ref{chapter:perturbation_analysis}: Operator-Theoretic Perturbation Analysis}, which is based on an article:
        \begin{enumerate}[resume=CompiledPublications,label=(\roman*)]
            \item \textcolor{Emerald}{\fullcite{luo2025spectralvariation}}.
        \end{enumerate}

        \item \textbf{Stochastic Dynamic Analysis of Lipschitz Continuity} (\hyperref[chapter:optimization_induced_dynamics]{Chapter~\ref{chapter:optimization_induced_dynamics}: Stochastic Dynamic Analysis of Lipschitz Continuity}) --- how does Lipschitz continuity evolve over time during training in neural networks? This is answered in \hyperref[chapter:optimization_induced_dynamics]{Chapter~\ref{chapter:optimization_induced_dynamics}: Stochastic Dynamic Analysis of Lipschitz Continuity}, which is based on an article:
        \begin{enumerate}[resume=CompiledPublications,label=(\roman*)]
            \item \textcolor{Emerald}{\fullcite{luo2025lipschitz}}.
        \end{enumerate}
    
\end{enumerate}

\paragraph{(RQ3) Modulation of Frequency Signal Propagation.} How does Lipschitz continuity modulate the frequency signal propagation in neural networks, and how does it help us for understanding the global robustness of image models? 
\begin{enumerate}[secondary-research-questions,resume=SecondaryResearchQuestions,left=2em]
        \item \textbf{Fourier Analysis of Lipschitz Continuity} (Chapter~\ref{chapter:fourier_analysis_of_lipschitz_continuity}: Fourier Analysis of Lipschitz Continuity) --- how does Lipschitz continuity modulate frequency signal propagation in image models? This is answered in \hyperref[chapter:fourier_analysis_of_lipschitz_continuity]{Chapter~\ref{chapter:fourier_analysis_of_lipschitz_continuity}: Fourier Analysis of Lipschitz Continuity}. 
        
        \item \textbf{Interpreting Global Robustness} (\hyperref[chapter:interpreting_global_robustness]{Chapter~\ref{chapter:interpreting_global_robustness}: Interpreting Global Robustness}) --- how can global robustness be explained through the lens of spectral bias in Lipschitz continuity? This is answered in \hyperref[chapter:interpreting_global_robustness]{Chapter~\ref{chapter:interpreting_global_robustness}: Interpreting Global Robustness}, which is based on an article:
        \begin{enumerate}[resume=CompiledPublications,label=(\roman*)]
            \item \textcolor{Emerald}{\fullcite{luo2024interpreting}}.
        \end{enumerate}
\end{enumerate}

\section{Contributions}

The main contributions are summarized as follows:

\begin{enumerate}[label=(\roman*)]
    \item \textbf{Comprehensive literature study.} We present an extensive survey on Lipschitz continuity in deep learning, covering theoretical foundations, estimation methods, regularization approaches, and certifiable robustness. This survey serves to consolidate existing knowledge.

    \item \textbf{Operator-Theoretic Perturbation Analysis.} We develop a formal operator-theoretic framework for analyzing spectral variations of parameter matrices in training. This framework enables precise characterization of parameter matrices during training.

    \item \textbf{Stochastic analysis of Lipschitz dynamics.} We introduce a theoretical framework that characterizes the temporal evolution of Lipschitz continuity of neural networks during training. This framework is formulated using a system of stochastic differential equations (SDEs) and enables a rigorous analysis of how Lipschitz continuity evolves throughout training. Our analysis identifies key factors that govern the temporal evolution of Lipschitz continuity of neural networks.

    \item \textbf{Interpreting Global Robustness.} We begin by analyzing Lipschitz continuity through the lens of Fourier analysis, revealing how it modulates frequency signal propagation in neural networks. Building on this foundation, we introduce a novel perspective for interpreting the global robustness of image models --- incorporating information theory and Shapley value theory --- by linking the spectral bias of Lipschitz continuity to robustness against input perturbations.
    
\end{enumerate}

}


\backmatter
\clearpage
\addcontentsline{toc}{chapter}{Bibliography}


\makeatletter
  \def\thesisbibfiles{%
    back_matter/common_references,%
    back_matter/compiled_publications,%
    back_matter/noncompiled_publications%
  }%
  \foreach \folder in \chapterfolders {%
    \xdef\thesisbibfiles{%
      \thesisbibfiles,%
      main_matter/chapters/\folder/references%
    }%
  }%

\makeatother

\bibliography{\thesisbibfiles}

\end{document}


\chapter{Stochastic Dynamic Analysis of Lipschitz Continuity}
\label{chapter:optimization_induced_dynamics}

{
\noindent\textbf{Copyright Disclaimer}. This chapter includes content adapted from the following article:
\begin{enumerate}[itemsep=0em,topsep=0em]
  \item \fullcite{luo2025lipschitz}
\end{enumerate}
\noindent Portions of the content are reused verbatim or with minor edits, in accordance with institutional guidelines and the policies of the relevant publishers. Permission from all co-authors has been obtained.

\noindent\textbf{Authorship Statement}. This chapter is the result of collaborative research. Contributions from the authors are stated as below:
\begin{enumerate}[itemsep=0em,topsep=0em,label=(\roman*)]
  \item R\'ois\'in Luo, affiliated with \textit{University of Galway -- Ireland} and \textit{Irish National Centre for Research Training in AI}, made the primary contributions, including conceptualization, theoretical analysis, drafting, \etc.

  \item Dr. James McDermott, affiliated with \textit{University of Galway -- Ireland} and \textit{Irish National Centre for Research Training in AI},  provided supervision, contributed to the refinement of the manuscript, offered academic guidance, and critically reviewed the content.

  \item Dr. Christian Gagn\'e, affiliated with \textit{Universit\'e Laval -- Canada}, \textit{Canada -- CIFAR AI Chair}, and \textit{Mila -- Qu\'ebec AI Institute}, contributed to the refinement of the manuscript, offered academic guidance, and critically reviewed the content.

  \item Dr. Qiang Sun, affiliated with \textit{University of Toronto - Canada}, and \textit{MBZUAI -- UAE}, contributed to the refinement of the manuscript, offered academic guidance, and critically reviewed the content.

  \item Dr. Colm O'Riordan, affiliated with \textit{University of Galway -- Ireland} and \textit{Irish National Centre for Research Training in AI},  provided supervision, contributed to the refinement of the manuscript, offered academic guidance, and critically reviewed the content.
  
\end{enumerate}
I confirm that the contributions stated above accurately reflect the work I completed and the assistance provided by co-authors. All collaborators have reviewed, approved the manuscript to which they contributed, and consent to its inclusion in this thesis.

}

\chapterline

{
\noindent\textbf{Guiding Research Question.} This chapter addresses the secondary research question \textbf{(SRQ4) Stochastic Dynamic Analysis of Lipschitz Continuity}, in service of the primary research question --- \textbf{(RQ2) Training Dynamics}. Building on the analytic operator-norm Hessian formulation introduced in the previous chapter, and incorporating high-dimensional stochastic differential equations (SDEs), this chapter establishes --- to the best of our knowledge --- the first theoretical framework for modeling the temporal evolution of Lipschitz continuity in neural networks during training. The proposed framework is validated through numerical experiments with a convolutional neural network trained on CIFAR-10 and CIFAR-100 with multiple regularization configurations. It also offers theoretical insights into neural network behaviors, including near convergence dynamics, the role of hyper-parameters, and the impact of noisy supervision.

\noindent\textbf{Reproducibility.}~~\url{https://anonymous.4open.science/r/lipschitz_dynamics_reproducibility-4721}

}

\begin{chapterabstract}
Lipschitz continuity characterizes the worst-case sensitivity of neural networks to small input perturbations; yet its dynamics (\ie~temporal evolution) during training remains under-explored. We present a rigorous mathematical framework to model the temporal evolution of Lipschitz continuity during training with stochastic gradient descent (SGD). This framework leverages a system of stochastic differential equations (SDEs) to capture both deterministic and stochastic forces. Our theoretical analysis identifies three principal factors driving the evolution: (i) the projection of gradient flows, induced by the optimization dynamics, onto the operator-norm Jacobian of parameter matrices; (ii) the projection of gradient noise, arising from the randomness in mini-batch sampling, onto the operator-norm Jacobian; and (iii) the projection of the gradient noise onto the operator-norm Hessian of parameter matrices. Furthermore, our theoretical framework sheds light on how noisy supervision, parameter initialization, batch size, and mini‐batch sampling trajectories, among other factors, shape the evolution of the Lipschitz continuity of neural networks. Our experimental results demonstrate strong agreement between the theoretical implications and the observed behaviors. 
\end{chapterabstract}


\lettrine[lines=3]{D}{eep} learning \citep{goodfellow2016deep} has achieved remarkable success across a diversity of domains, including computer vision \citep{krizhevsky2012imagenet,he2016deep,dosovitskiy2020image}, natural language processing \citep{hochreiter1997long,vaswani2017attention,radford2019language,devlin2019bert,brown2020language,openai2023gpt4,deepseekai2025deepseekr1}, and graph representation learning \citep{perozzi2014deepwalk,velivckovic2018graph,xu2018powerful}. These successes have largely expanded the frontiers of what is achievable in artificial intelligence, enabling powerful capabilities across a wide range of tasks, as well as modeling paradigms such as generative modeling \citep{hinton2006reducing,goodfellow2014generative,dinh2017density,ho2020denoising,song2020score,lipman2023flow}. 

Despite these remarkable advances, critical challenges persist, notably in ensuring \textbf{robustness} to small input perturbations and \textbf{generalization} to out-of-distribution data \citep{szegedy2014intriguing,goodfellow2014explaining,hendrycks2019benchmarking,ilyas2019adversarial,recht2019imagenet,luo2024interpreting}. These critical challenges highlight the need to understand the underlying fundamental principles that govern \textbf{robustness} and \textbf{generalization}, and to establish deep learning systems that are \textbf{reliable} --- performing consistently under expected conditions on in-distribution data \citep{nist2023ai}, \textbf{resilient} --- capable of recovering from unexpected conditions such as noise or adversarial attacks \citep{nist2023ai}, and \textbf{trustworthy} --- behaving transparently, ethically, in alignment with intended use, and technically robust \citep{eu-trustworthy-ai-guidelines,eu-ai-act-2024} --- in safety-critical applications.

\begin{figure}[t!]
    \centering
    
    \resizebox{1\linewidth}{!}{
      \begin{tikzpicture}[
  node distance=2cm,
  databox/.style={
    cylinder,
    shape border rotate=90,
    draw,
    minimum width=2cm,
    minimum height=2.5cm
  },
  funcbox/.style={
    rectangle,
    draw,
    thick,
    minimum width=8cm,
    minimum height=4.5cm
  },
  ellipseblock/.style={
    ellipse,
    draw,
    thick,
    minimum width=2.5cm,
    minimum height=2.5cm
  },
  circleblock/.style={
    circle,
    draw,
    thick,
    minimum size=2cm
  },
  arrow/.style={
        -{Classical TikZ Rightarrow[length=3mm]},
        thick
    }
]

  %
  \makeatletter
  \newcommand{\gear}[5]{%
    \foreach \i in {1,...,#1}{%
      [rotate=(\i-1)*360/#1]
        (0:#2) arc (0:#4:#2)
        -- (#4+#5:#3) arc (#4+#5:360/#1-#5:#3)
        -- (360/#1:#2)
    }%
  }
  \makeatother
  %

  \node[databox]                   (data)      {Data: $x \in \mathcal{X}$};
  
  \node[below=0.3cm of data, text width=8cm, align=center,text=Emerald] (lipschitztext) {\textbf{Lipschitz continuity modulates data feature propagation in $f$.}};
  
  \node[funcbox, right=of data] (function) {}; 
  \node at ([xshift=0cm, yshift=1.4cm]function.center) {Learned function: $f(x;\boldsymbol{\theta}): \mathcal{X} \times \Theta \to \tilde{\mathcal{Y}}$};

    \node[
  anchor=north,
  text width=10cm,
  align=center,text=Emerald
    ] at ([yshift=-0.3cm]function.south) 
    {\textbf{Optimization induces the dynamics of Lipschitz continuity in $f$.}};

  \node[ellipseblock, above=of function]     (prior)     {Supervision signal: $y \in \mathcal{Y}$};
  \node[circleblock,  right=of function] (criterion) {Criterion: $\ell_f(f(x;\boldsymbol{\theta}),y)$};

  \draw[arrow] (data)     -- (function);
  \draw[arrow] (function) -- (criterion);
  \draw[arrow] ([yshift=-1cm]data.north)     |- (prior);
  \draw[arrow] (prior)    -| (criterion);

  \begin{scope}[shift={(function.center)}]
    \def\nteeth{18}
    \def\rin{0.8}     
    \def\rout{1}      
    \def\toothang{10} 
    \def\gapang{2}    
    \def\halfPitch{180/\nteeth}

    \coordinate (G1) at (-2*\rout,0);
    \coordinate (G2) at (   0,   -1cm);
    \coordinate (G3) at ( 2*\rout,0);

    \begin{scope}[fill=gray!50, thick]
      \draw[shift={(G1)}] \gear{\nteeth}{\rin}{\rout}{\toothang}{\gapang};
      \draw[shift={(G2)},rotate=\halfPitch]
        \gear{\nteeth}{\rin}{\rout}{\toothang}{\gapang};
      \draw[shift={(G3)}] \gear{\nteeth}{\rin}{\rout}{\toothang}{\gapang};
    \end{scope}

    \node at (G1) {$\boldsymbol{\theta}^{(1)}$};
    \node at (G2) {$\boldsymbol{\theta}^{(2)}$};
    \node at (G3) {$\boldsymbol{\theta}^{(3)}$};
  \end{scope}

\end{tikzpicture}
    }

    \caption[Principles of Lipschitz Continuity Manifest in Paradigm of Machine Learning]{\textbf{Principles of Lipschitz Continuity Manifest in Paradigm of Machine Learning}. In the paradigm of machine learning, we consider \textit{data} $x \in \mathcal{X}$, \textit{supervision signal} $y \in \mathcal{Y}$, a \textit{learned function} $f: \mathcal{X} \times \boldsymbol{\Theta} \to \Tilde{\mathcal{Y}}$ --- where $\Tilde{\mathcal{Y}}$ is output space --- parameterized by $\theta^{(\ell)} \in \mathbb{R}^{d_\ell}$ and $\boldsymbol{\Theta}:=\mathbb{R}^{d_1 \times d_2 \times \cdots \times d_L}$, and a \textit{criterion} $\ell_f: (f(x, \boldsymbol{\theta}), \mathcal{Y}) \to \mathbb{R}$. The Lipschitz constant $K_f$ of $f$ in this paradigm is determined by factors such as the hypothesis class for $f$, learning algorithm, parameters $\boldsymbol{\theta}$, training data $(\mathcal{X}, \mathcal{Y})$, \etc. It also modulates the behavior of $f$ with respect to the input features. This thesis investigates the \textbf{principles of Lipschitz continuity} in this paradigm from a \textbf{complementary dichotomy of perspectives}: an \textbf{internal} perspective --- focusing on the temporal evolution of Lipschitz continuity in neural networks during training; and an \textbf{external} perspective --- examining how Lipschitz continuity modulates the behavior of $f$ with respect to the features in input data, particularly its role in governing frequency signal propagation in neural networks.}

    \label{fig:paradigm_of_machine_learning}

\end{figure}

The paradigm of machine learning \citep{goodfellow2016deep}, illustrated in Figure~\ref{fig:paradigm_of_machine_learning}, typically comprises the \emph{data} $x \in \mathcal{X} \subseteq \mathbb{R}^m$, \emph{supervision signal} $y \in \mathcal{Y} $, a \emph{learned function} $f(x; \boldsymbol{\theta})$ with output space $\Tilde{\mathcal{Y}} \subseteq \mathbb{R}^n$:
\begin{align}
    f: \mathcal{X} \times \boldsymbol{\Theta} \to \Tilde{\mathcal{Y}}
    ,
\end{align}
parameterized by:
\begin{align}
\boldsymbol{\theta}:=\{\boldsymbol{\theta}^{(1)}, \boldsymbol{\theta}^{(2)},\cdots \} \in \boldsymbol{\Theta}
\subseteq \mathbb{R}^p
,
\end{align}
and a \emph{criterion}:
\begin{align}
    \ell_f: (f(x; \boldsymbol{\theta}), y) \to \mathbb{R}.
\end{align}
Note that the \emph{supervision signals} differ across learning paradigms, for example:
\begin{enumerate}[label=(\roman*)]
    \item \textbf{Supervised Learning.} The \emph{supervision signals} are explicitly provided as \emph{labels} by the dataset;

    \item \textbf{Self-Supervised Learning.} The \emph{supervision signals} are derived from inherent structure or priors in the data --- for example, in contrastive learning, an augmented view of an image can serve as the label for the original image \citep{chen2020simple};

    \item \textbf{Reinforcement Learning.} The \emph{supervision signals} are manifested as \emph{reward signals} from the interactions with the environment, which are often noisy, sparse, and delayed \citep[Preface \& \S~1]{sutton1998reinforcement}.

\end{enumerate}

Among the theoretical tools available, \textbf{Lipschitz continuity} \citep[Theorem 9.19, \S~9]{rudin1976principles} plays a pivotal role in governing the fundamental properties of the function $f$ in this paradigm related to robustness and generalization. The \emph{Lipschitz constant} $K_f$ of $f$, with respect to the $\ell_2$-norm, defined as:
\begin{align}
    K_f := \sup_{\forall~u \neq v} \frac{\|f(u) - f(v)\|_2}{\|u - v\|_2},
\end{align}
measures the worst-case sensitivity of the network to small input perturbations. Equivalently, it bounds the maximum rate of change of $f$ with respect to its inputs in the $\ell_2$-norm \citep[Theorem 9.19, \S~9]{rudin1976principles}:
\begin{align}
K_f \geq \sup_x \|\nabla_x f(x) \|_2
    ,
\end{align}
where $\nabla_x (\cdot)$ denotes a differential operator acting on $(\cdot)$ with respect to $x$. A lower Lipschitz constant $K_f$ ensures a smoother mapping $f$, enhancing the network's ability to generalize to unseen data. Moreover, Lipschitz continuity forms a theoretical foundation for \textbf{certifiable robustness} --- where safety guarantees under bounded perturbations can be formally proven --- in safety-critical applications. 

While Lipschitz continuity is crucial in governing the fundamental properties of neural networks related to robustness and generalization, prior research has largely concentrated on empirical aspects, such as regularization techniques \citep{miyato2018spectral,arjovsky2017wasserstein,cisse2017parseval,anil2019sorting}, leaving the underlying principles less explored. This thesis is motivated by the absence of theoretical underpinnings of the principles of Lipschitz continuity in neural networks --- a foundation that is critical for establishing reliable, resilient, and trustworthy deep learning systems. To address this gap, it seeks to advance a principled understanding of the \textbf{principles of Lipschitz continuity in neural networks} within the paradigm of machine learning, examined from \textbf{a complementary dichotomy of perspectives}: 
\begin{enumerate}[label={}]
    \item an \textbf{internal} perspective --- focusing on the temporal evolution of Lipschitz continuity in neural networks during training (\ie, \textbf{Training Dynamics});

    \item an \textbf{external} perspective --- investigating how Lipschitz continuity modulates the behavior of neural networks with respect to features in the input data, particularly its role in governing frequency signal propagation (\ie, \textbf{Modulation of Frequency Signal Propagation}).
\end{enumerate}
Together, these complementary perspectives form the basis of the primary research questions presented in Section~\ref{sec:research_questions} and underscore the principles under investigation within this thesis.

\section{Research questions}
\label{sec:research_questions}


This thesis is centered around an overarching inquiry from a complementary dichotomy of \textbf{internal} and \textbf{external} perspectives:
\begin{quote}
\textit{What are the \textbf{principles of Lipschitz continuity in neural networks}, from its temporal evolution during training (an \textbf{internal perspective}) to its modulation of frequency signal propagation (an \textbf{external perspective})?}
\end{quote}

To support this central inquiry, it investigates three primary research questions (RQs) with a structured approach, each gives rise to a set of secondary research questions (SRQs), which are explored in one or two dedicated chapters.

\begin{questionbox}{(RQ1) State of Knowledge}
\begin{treelist}
    \item (RQ1) \textbf{State of Knowledge} --- what is the state of knowledge of Lipschitz continuity in deep learning? 
    
    \begin{treelist}
        \item (SRQ1) \textbf{Theoretical Foundations} --- what are the theoretical foundations of Lipschitz continuity in deep learning? 
        
        \item (SRQ2) \textbf{Regularization Mechanism} --- how does the constraint of Lipschitz continuity serve as a regularization mechanism for improving robustness and generalization capabilities in deep learning?
        
    \end{treelist}
\end{treelist}
\end{questionbox}

\begin{questionbox}{(RQ2) Training Dynamics --- An Internal Perspective}
\begin{treelist}
    \item (RQ2) \textbf{Training Dynamics} --- how does Lipschitz continuity of neural networks evolve over time during training?
    \begin{treelist}
        \item (SRQ3) \textbf{Parameter Space Perturbation} --- how do the spectral properties (\eg~singular values) of parameter matrices change in training? During training, the optimization process updates parameter matrices, thereby altering the spectral structures accordingly. 

        \item (SRQ4) \textbf{Stochastic Dynamic Analysis of Lipschitz Continuity} --- how does Lipschitz continuity evolve over time during training in neural networks? 
        
    \end{treelist}
\end{treelist}
\end{questionbox}

\begin{questionbox}{(RQ3) Modulation of Frequency Signal Propagation --- An External Perspective}
\begin{treelist}
    \item (RQ3) \textbf{Modulation of Frequency Signal Propagation} --- how does Lipschitz continuity modulate the frequency signal propagation in neural networks, and how does it help us for understanding the global robustness of image models?
    \begin{treelist}
        \item (SRQ5) \textbf{Fourier Analysis of Lipschitz Continuity} --- how does Lipschitz continuity modulate frequency signal propagation in image models?
        
        \item (SRQ6) \textbf{Interpreting Global Robustness} --- how can global robustness be explained through the lens of spectral bias in Lipschitz continuity? 
    \end{treelist}
\end{treelist}
\end{questionbox}

\section{Thesis Presentation Format}

This thesis compiles and presents a selection of research articles --- published or under review as of the date of submission --- into a \underline{\textbf{self-contained}}, \underline{\textbf{modular}}, \underline{\textbf{coherent}}, \underline{\textbf{monograph-style}} presentation format, all centered on the principles of Lipschitz continuity in neural networks. The research was conducted cumulatively during the Ph.D. training. Portions of the chapter content, which address the research questions, are reused either verbatim or with minor modifications from the selected articles, in accordance with institutional guidelines and the policies of the relevant publishers. Each core chapter is presented in the style of a standalone journal article, including an \textbf{abstract}, \textbf{introduction}, and other standard components in journal articles. For consistency with the original publications, the term \textbf{paper} is used throughout to refer to the current chapter, and the literature review may include results from our research, either published or available as preprints. Permissions for reuses have been obtained from all co-authors in all articles.

\section{Content Organization}

\begin{figure}[t!]
    \centering
    
    \resizebox{1\linewidth}{!}{





\begin{tikzpicture}[node distance=10mm and 15mm, >=Stealth,
  rq1/.style={
        rectangle, draw, rounded corners, 
        fill=green!20, 
        align=center, text width=5cm, minimum height=2cm},
  rq2/.style={
        rectangle, draw, rounded corners, 
        fill=orange!20, 
        align=center, text width=5cm, minimum height=2cm},
  rq3/.style={
        rectangle, draw, rounded corners, 
        fill=cyan!20, 
        align=center, text width=5cm, minimum height=2cm},
  legendbox/.style={
        rectangle, draw, rounded corners, minimum width=12mm, minimum height=12mm, inner sep=0pt},
  legend label/.style={
        anchor=west, 
        }
]

  \begin{scope}[local bounding box=diagram]
  
    \node[rq1] (LiteratureReview) {\hyperref[chapter:literature_review]{Chapter~\ref{chapter:literature_review}: Literature Review}};

    \node[rq1, ellipse, draw, align=center, text width=3.5cm, minimum height=2cm, right=of LiteratureReview, yshift=15mm] (TheoreticalFoundations) {Theoretical Foundations};

    \node[rq1, ellipse, draw, align=center, text width=3.5cm, minimum height=2cm, right=of LiteratureReview, yshift=-15mm] (Regularizations) {Regularizations};

    \node[rq2, above=of TheoreticalFoundations] (PerturbationAnalysis) {\hyperref[chapter:perturbation_analysis]{Chapter~\ref{chapter:perturbation_analysis}: Operator-Theoretic Perturbation Analysis}};

    \node[rq2, ellipse, draw, align=center, text width=3.5cm, minimum height=2cm, left=of PerturbationAnalysis, yshift=15mm] (PerturbationTheory) {Perturbation Theory};

    \node[rq2, ellipse, draw, align=center, text width=3.5cm, minimum height=2cm, left=of PerturbationAnalysis, yshift=-15mm] (OperatorTheory) {Operator Theory};

    \node[rq2, right=of TheoreticalFoundations] (InducedDynamics) {\hyperref[chapter:optimization_induced_dynamics]{Chapter~\ref{chapter:optimization_induced_dynamics}: Stochastic Dynamic Analysis of Lipschitz Continuity}};

    \node[rq2, ellipse, draw, align=center, text width=3.5cm, minimum height=2cm, above=of InducedDynamics] (SDEFoundations) {Matrix-Valued SDE};

    \node[rq2, ellipse, draw, align=center, text width=3.5cm, minimum height=2cm, below=of InducedDynamics] (SGDOptimization) {SGD Optimization};

    \node[rq2, ellipse, draw, align=center, text width=3.5cm, minimum height=2cm, right=of InducedDynamics] (TheoreticalImplications) {Theoretical Implications in Training Dynamics};

    \node[rq2, ellipse, draw, align=center, text width=3.5cm, minimum height=2cm, right=of TheoreticalImplications] (NoisyGradientRegularizations) {Noisy Gradient Regularizations};

    \node[rq2, ellipse, draw, align=center, text width=3.5cm, minimum height=2cm, above=of NoisyGradientRegularizations] (NearConvergenceBehaviours) {Near Convergence Behaviours};

    \node[rq2, ellipse, draw, align=center, text width=3.5cm, minimum height=2cm, above=of NearConvergenceBehaviours] (ParameterInitialization) {Robustness due to Parameter Initialization};

    \node[rq2, ellipse, draw, align=center, text width=3.5cm, minimum height=2cm, below=of NoisyGradientRegularizations] (BatchSize) {Batch Size};

    \node[rq2, ellipse, draw, align=center, text width=3.5cm, minimum height=2cm, below=of BatchSize] (SamplingTrajectory) {Mini-Batch Sampling Trajectory};
    
    \node[rq3, below=of Regularizations] (LocalLipschitz) {\hyperref[chapter:fourier_analysis_of_lipschitz_continuity]{Chapter~\ref{chapter:fourier_analysis_of_lipschitz_continuity}: Fourier Analysis of Lipschitz Continuity}};

    \node[rq3, right=of LocalLipschitz] (GlobalInterpretation) {\hyperref[chapter:interpreting_global_robustness]{Chapter~\ref{chapter:interpreting_global_robustness}: Interpreting Global Robustness}};

    \node[rq3, ellipse, draw, align=center, text width=3.5cm, minimum height=2cm, below=of GlobalInterpretation, xshift=-30mm] (GameTheory) {Shapley Value Theory};

    \node[rq3, ellipse, draw, align=center, text width=3.5cm, minimum height=2cm, below=of GlobalInterpretation, xshift=30mm] (InformationTheory) {Information Theory};

    \draw[->] (LiteratureReview) -- (TheoreticalFoundations);
    \draw[->] (LiteratureReview) -- (Regularizations);
    \draw[->] (OperatorTheory) -- (PerturbationAnalysis);
    \draw[->] (PerturbationTheory) -- (PerturbationAnalysis);
    \draw[->] (TheoreticalFoundations) -- (InducedDynamics);
    \draw[->] (SGDOptimization) -- (InducedDynamics);
    \draw[->] (SDEFoundations) -- (InducedDynamics);
    \draw[->] (PerturbationAnalysis) -- (InducedDynamics);
    \draw[->] (InducedDynamics) -- (TheoreticalImplications);
    
    \draw[->] (TheoreticalImplications) -- (NoisyGradientRegularizations);
    \draw[->] (TheoreticalImplications) -- (NearConvergenceBehaviours);
    \draw[->] (TheoreticalImplications) -- (ParameterInitialization);
    \draw[->] (TheoreticalImplications) -- (BatchSize);
    \draw[->] (TheoreticalImplications) -- (SamplingTrajectory);
    
    \draw[->] (Regularizations) -- (LocalLipschitz);
    \draw[->] (LocalLipschitz) -- (GlobalInterpretation);
    \draw[->] (GameTheory) -- (GlobalInterpretation);
    \draw[->] (InformationTheory) -- (GlobalInterpretation);

    \path (current bounding box.north east) coordinate (bbne);
    \path (current bounding box.south west) coordinate (bbsw);
    \path[use as bounding box] (bbsw) rectangle (bbne);
    
  \end{scope}

  \begin{scope}[local bounding box=legend, 
                above=of diagram,
                anchor=south,
                xshift=50mm,
                yshift=90mm
                ]
    \node[legendbox, fill=green!20] (legC) {};
    \node[legend label, right=1mm of legC] (labC) {RQ1: \textbf{State of Knowledge}};
    
    \node[legendbox, fill=orange!20, right=5cm of legC] (legO) {};
    \node[legend label, right=1mm of legO] (labO) {RQ2: \textbf{Training Dynamics}};
    
    \node[legendbox, fill=cyan!20, right=5cm of legO] (legG) {};
    \node[legend label, right=1mm of legG] (labG) {RQ3: \textbf{Modulation of Frequency Signal Propagation}};
  \end{scope}
  
\end{tikzpicture}

    }
    
    \caption[Thesis Thematic Structure]{\textbf{Thesis Thematic Structure}. A high-level conceptual diagram outlining the thematic connections between chapters. This thesis investigates the principles of Lipschitz continuity in neural networks through a structured approach guided by three primary research questions. Each primary question gives rise to two secondary research questions and is addressed in one or two dedicated chapters.}

    \label{fig:thesis_thematic_structure}
\end{figure}


The research questions collectively establish a cohesive thematic investigation into the principles of Lipschitz continuity in neural networks. Figure~\ref{fig:thesis_thematic_structure} illustrates how each question aligns with the core chapters and jointly guides the development of the overarching theoretical framework. The structure of the thesis is outlined below.

\paragraph{(RQ1) State of Knowledge.} What is the state of knowledge of Lipschitz continuity in deep learning? This research question is addressed by surveying the landscape of Lipschitz continuity in deep learning, covering its theoretical foundations, estimation methods, regularization approaches, and role in certifiable robustness. This research question is addressed by surveying the landscape of Lipschitz continuity in deep learning, covering its theoretical foundations, estimation methods, regularization approaches, and role in certifiable robustness. 
\begin{enumerate}[secondary-research-questions,series=SecondaryResearchQuestions,left=2em]
        \item \textbf{Theoretical Foundations} (\hyperref[chapter:literature_review]{Chapter~\ref{chapter:literature_review}: Literature Review}) --- what are the theoretical foundations of Lipschitz continuity in deep learning? This is answered in \hyperref[chapter:literature_review]{Chapter~\ref{chapter:literature_review}: Literature Review}, which is based on an article:
        \begin{enumerate}[label=(\roman*),series=CompiledPublications]
            \item \textcolor{Emerald}{\fullcite{luo2025lipschitzsurvey}}.
        \end{enumerate}
    
        \item \textbf{Regularization Mechanism} (\hyperref[chapter:literature_review]{Chapter~\ref{chapter:literature_review}: Literature Review}) --- how does Lipschitz continuity serve as a regularization mechanism for improving robustness and generalization capabilities in deep learning? This is answered in \hyperref[chapter:literature_review]{Chapter~\ref{chapter:literature_review}: Literature Review}, which is based on an article:
        \begin{enumerate}[label=(\roman*),series=CompiledPublications]
            \item \textcolor{Emerald}{\fullcite{luo2025lipschitzsurvey}}.
        \end{enumerate}
    \end{enumerate}

\paragraph{(RQ2) Training Dynamics.} How does Lipschitz continuity evolve over time during the training in neural networks? To the best of our knowledge, as of the date of submission, this constitutes the first theoretical framework to characterize the temporal evolution of Lipschitz continuity in neural networks leveraging perturbation theory \citep{kato1995perturbation,luo2025spectralvariation} and a system of stochastic differential equations (SDEs). 
\begin{enumerate}[secondary-research-questions,resume=SecondaryResearchQuestions,left=2em]
        \item \textbf{Parameter Space Perturbation} (\hyperref[chapter:perturbation_analysis]{Chapter~\ref{chapter:perturbation_analysis}: Operator-Theoretic Perturbation Analysis}) --- how do the spectral properties (\ie~singular values) of parameter matrices change in training? During training, the optimization process updates parameter matrices, thereby altering the spectral structures accordingly. This is addressed in \hyperref[chapter:perturbation_analysis]{Chapter~\ref{chapter:perturbation_analysis}: Operator-Theoretic Perturbation Analysis}, which is based on an article:
        \begin{enumerate}[resume=CompiledPublications,label=(\roman*)]
            \item \textcolor{Emerald}{\fullcite{luo2025spectralvariation}}.
        \end{enumerate}

        \item \textbf{Stochastic Dynamic Analysis of Lipschitz Continuity} (\hyperref[chapter:optimization_induced_dynamics]{Chapter~\ref{chapter:optimization_induced_dynamics}: Stochastic Dynamic Analysis of Lipschitz Continuity}) --- how does Lipschitz continuity evolve over time during training in neural networks? This is answered in \hyperref[chapter:optimization_induced_dynamics]{Chapter~\ref{chapter:optimization_induced_dynamics}: Stochastic Dynamic Analysis of Lipschitz Continuity}, which is based on an article:
        \begin{enumerate}[resume=CompiledPublications,label=(\roman*)]
            \item \textcolor{Emerald}{\fullcite{luo2025lipschitz}}.
        \end{enumerate}
    
\end{enumerate}

\paragraph{(RQ3) Modulation of Frequency Signal Propagation.} How does Lipschitz continuity modulate the frequency signal propagation in neural networks, and how does it help us for understanding the global robustness of image models? 
\begin{enumerate}[secondary-research-questions,resume=SecondaryResearchQuestions,left=2em]
        \item \textbf{Fourier Analysis of Lipschitz Continuity} (Chapter~\ref{chapter:fourier_analysis_of_lipschitz_continuity}: Fourier Analysis of Lipschitz Continuity) --- how does Lipschitz continuity modulate frequency signal propagation in image models? This is answered in \hyperref[chapter:fourier_analysis_of_lipschitz_continuity]{Chapter~\ref{chapter:fourier_analysis_of_lipschitz_continuity}: Fourier Analysis of Lipschitz Continuity}. 
        
        \item \textbf{Interpreting Global Robustness} (\hyperref[chapter:interpreting_global_robustness]{Chapter~\ref{chapter:interpreting_global_robustness}: Interpreting Global Robustness}) --- how can global robustness be explained through the lens of spectral bias in Lipschitz continuity? This is answered in \hyperref[chapter:interpreting_global_robustness]{Chapter~\ref{chapter:interpreting_global_robustness}: Interpreting Global Robustness}, which is based on an article:
        \begin{enumerate}[resume=CompiledPublications,label=(\roman*)]
            \item \textcolor{Emerald}{\fullcite{luo2024interpreting}}.
        \end{enumerate}
\end{enumerate}

\section{Contributions}

The main contributions are summarized as follows:

\begin{enumerate}[label=(\roman*)]
    \item \textbf{Comprehensive literature study.} We present an extensive survey on Lipschitz continuity in deep learning, covering theoretical foundations, estimation methods, regularization approaches, and certifiable robustness. This survey serves to consolidate existing knowledge.

    \item \textbf{Operator-Theoretic Perturbation Analysis.} We develop a formal operator-theoretic framework for analyzing spectral variations of parameter matrices in training. This framework enables precise characterization of parameter matrices during training.

    \item \textbf{Stochastic analysis of Lipschitz dynamics.} We introduce a theoretical framework that characterizes the temporal evolution of Lipschitz continuity of neural networks during training. This framework is formulated using a system of stochastic differential equations (SDEs) and enables a rigorous analysis of how Lipschitz continuity evolves throughout training. Our analysis identifies key factors that govern the temporal evolution of Lipschitz continuity of neural networks.

    \item \textbf{Interpreting Global Robustness.} We begin by analyzing Lipschitz continuity through the lens of Fourier analysis, revealing how it modulates frequency signal propagation in neural networks. Building on this foundation, we introduce a novel perspective for interpreting the global robustness of image models --- incorporating information theory and Shapley value theory --- by linking the spectral bias of Lipschitz continuity to robustness against input perturbations.
    
\end{enumerate}

\newpage
\begin{subappendices}
\section{Appendix}
\section{Network configuration}

\begin{table}[h]
    \centering
    \adjustbox{max width=\textwidth}{%
    \begin{tabular}{|c|l|l|}
        \hline
        
        \textbf{Function Block} &\textbf{\# of Layer} &  \textbf{Module} \\
        \hline
        
        \multirow{3}{*}{Block \#1} & \cellcolor{lime!20}$1$ & \cellcolor{lime!20}\texttt{nn.Conv2d(3, 32, kernel\_size=3, padding=1)} \\
        \cline{2-3}
        
        &$2$ & \texttt{nn.ReLU()} \\
        \cline{2-3}

        &$3$ & \texttt{nn.MaxPool2d(kernel\_size=2, stride=2)} \\
        \hline

        \multirow{4}{*}{Block \#2}& \cellcolor{lime!20}$4$ &\cellcolor{lime!20}\texttt{nn.Conv2d(32, 64, kernel\_size=3, padding=1)} \\
        \cline{2-3}

        &$5$ (*) & \texttt{nn.Dropout(p=0.2)} \\
        \cline{2-3}

        &$6$ & \texttt{nn.ReLU()} \\
        \cline{2-3}

        &$7$ & \texttt{nn.MaxPool2d(kernel\_size=2, stride=2)} \\
        \hline

        \multirow{4}{*}{Block \#3}& \cellcolor{lime!20}$8$ &\cellcolor{lime!20}\texttt{nn.Conv2d(64, 128, kernel\_size=3, padding=1, stride=2)} \\
        \cline{2-3}

        &$9$ (*) & \texttt{nn.Dropout(p=0.2)} \\
        \cline{2-3}

        &$10$ (*) & \texttt{nn.BatchNorm2d(128, track\_running\_stats=False)} \\
        \cline{2-3}

        &$11$& \texttt{nn.ReLU()} \\
        \hline

        \multirow{4}{*}{Classifier} & \cellcolor{lime!20}$12$ &\cellcolor{lime!20}\texttt{nn.Linear(128 * 4 * 4, 256)} \\
        \cline{2-3}

        &$13$ & \texttt{nn.ReLU()} \\
        \cline{2-3}

        &$14$ (*) & \texttt{nn.Dropout(p=0.3)} \\
        \cline{2-3}

        &\cellcolor{lime!20}$15$ & \cellcolor{lime!20}\texttt{nn.Linear(256, num\_classes)} \\
        \hline
        
    \end{tabular}
    }
    \caption{ConvNet configuration. The layer marked as `(*)' is configurable with respect to experimental needs. There are five parameterized layers (\ie~\# 1, \# 4, \# 8, \# 12 and \# 15).}
    
    \label{tab:network_config_cnn}
\end{table}

\begin{table}[h]
    \centering
    \begin{tabular}{|l|l|}
        \hline
        
        \textbf{\# of Layer} &  \textbf{Module} \\
        \hline
        
        \cellcolor{lime!20}$1$ & \cellcolor{lime!20}\texttt{nn.Linear(28*28, 512)} \\
        \hline
        
        $2$ & \texttt{nn.ReLU()} \\
        \hline

         \cellcolor{lime!20}$3$ & \cellcolor{lime!20}\texttt{nn.Linear(512, 256)} \\
        \hline

        $4$ & \texttt{nn.ReLU()} \\
        \hline

        \cellcolor{lime!20}$5$ & \cellcolor{lime!20}\texttt{nn.Linear(256, 10)} \\
        \hline
        
    \end{tabular}
    \caption{MLP configuration. There are three parameterized layers (\ie~\# 1, \# 3, and \# 5).}
    
    \label{tab:network_config_mlp}
\end{table}

\section{Regularization configuration}

\begin{table}[h]
    \centering
    \adjustbox{max width=\textwidth}{%
    \begin{tabular}{|c|l|}
        \hline
        
        \textbf{Regularization} & \textbf{Configuration} \\
        \hline

        mixup & $\alpha=0.4$ \\
        \hline

        label smoothing & $\epsilon=0.1$ \\
        \hline
        
        adversarial training w/ FGSM & $\epsilon=0.03$ \\
        \hline
  
        weight-decay & $\lambda=0.001$ \\
        \hline

        \multirow{2}{*}{auto-augment} & \texttt{AutoAugment(policy=AutoAugmentPolicy.CIFAR10)} \\
        \cline{2-2}
        
        & \texttt{AutoAugment(policy=AutoAugmentPolicy.CIFAR100)} \\
        \hline
        
    \end{tabular}
    }
    \caption{Regularization configuration.}
    
    \label{tab:regularization_config}
\end{table}

\section{Full validation experiment}
\label{sec:full_validation_cifar10}

\begin{sidewaysfigure}
  \centering

    \includegraphics[width=1\linewidth]{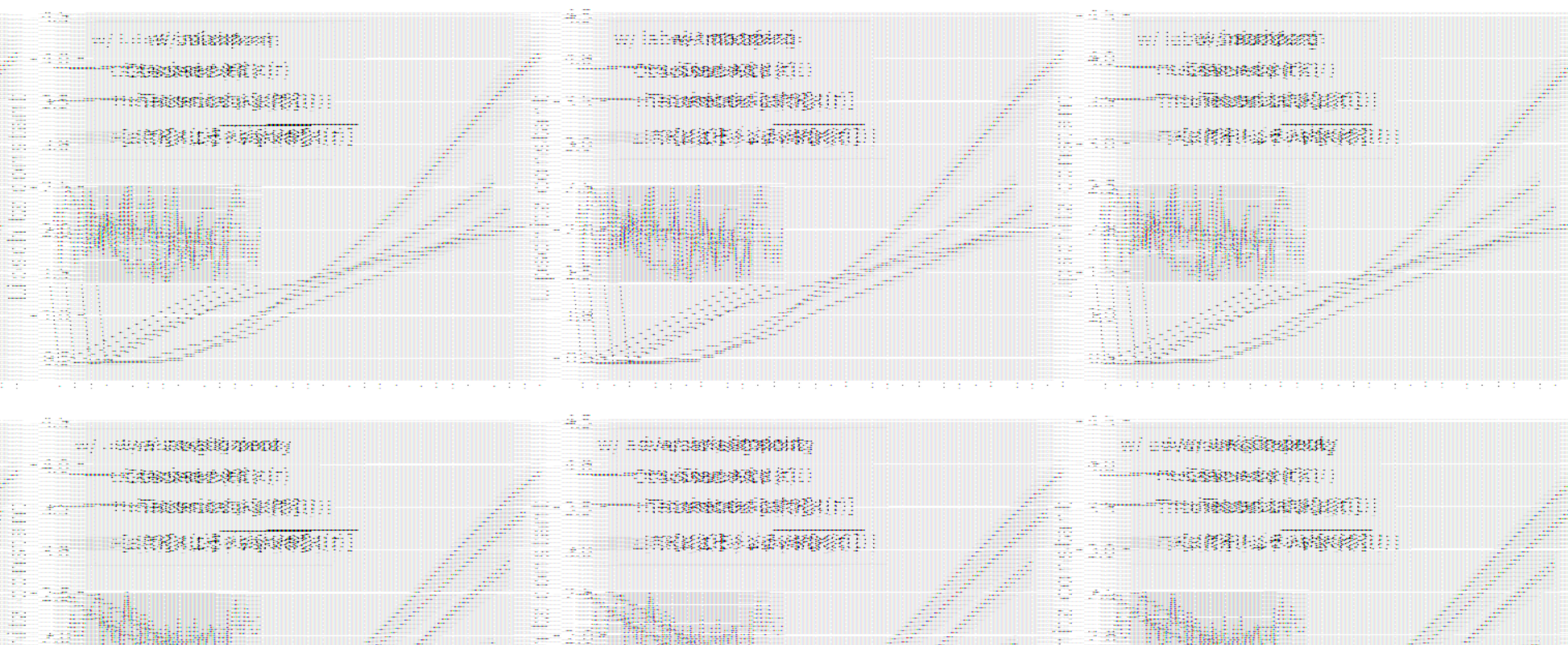}

   \caption[Full Numerical Validation of Our Mathematical Framework on CIFAR-10]{Full numerical validation of our mathematical framework on CIFAR-10.}
  
   \label{fig:validation:cifar10:full}
  
\end{sidewaysfigure}

Results of full validation experiments are shown as in Figure~\ref{fig:validation:cifar10:full}.

\end{subappendices}


\chapter{Interpreting Global Robustness}

\label{chapter:interpreting_global_robustness}

{
\noindent\textbf{Copyright Disclaimer}. This chapter includes content adapted from the following article:
\begin{enumerate}
  \setlength{\itemsep}{0pt} 
  \setlength{\parskip}{0pt}
  \item \fullcite{luo2024interpreting}
\end{enumerate}
\noindent Portions of the content are reused verbatim or with minor edits, in accordance with institutional guidelines and the policies of the relevant publishers. Permission from all co-authors has been obtained.
}

\chapterline

{
\noindent\textbf{Guiding Research Question.} This chapter addresses the secondary research question \textbf{(SRQ6) Interpreting Global Robustness}, in service of the primary research question --- \textbf{(RQ3) Modulation of Frequency Signal Propagation}. Building on the theoretical analysis and empirical study from the previous chapter, it presents a method for interpreting the global robustness of image models to input perturbations. The results further support and extend the theoretical and empirical insights established in the previous chapter.

%

\noindent\textbf{Reproducibility.}~~\url{https://anonymous.4open.science/r/lipschitz_dynamics_reproducibility-4721}
}

\begin{chapterabstract}
Perturbation robustness evaluates the vulnerabilities of models, arising from a variety of perturbations, such as data corruptions and adversarial attacks. Understanding the mechanisms of perturbation robustness is critical for global interpretability. We present a model-agnostic, global mechanistic interpretability method to interpret the perturbation robustness of image models. This research is motivated by two key aspects. First, previous global interpretability works, in tandem with robustness benchmarks, \eg~mean corruption error (mCE), are not designed to directly interpret the mechanisms of perturbation robustness within image models. Second, we notice that the spectral signal-to-noise ratios (SNR) of perturbed natural images exponentially decay over the frequency. This power-law-like decay implies that: Low-frequency signals are generally more robust than high-frequency signals --- yet high classification accuracy can not be achieved by low-frequency signals alone. By applying Shapley value theory, our method axiomatically quantifies the predictive powers of robust features and non-robust features within an information theory framework. Our method, dubbed as \textbf{I-ASIDE} (\textbf{I}mage \textbf{A}xiomatic \textbf{S}pectral \textbf{I}mportance \textbf{D}ecomposition \textbf{E}xplanation), provides a unique insight into model robustness mechanisms. We conduct extensive experiments over a variety of vision models pre-trained on ImageNet, including both convolutional neural networks (\eg~\textit{AlexNet}, \textit{VGG}, \textit{GoogLeNet/Inception-v1}, \textit{Inception-v3}, \textit{ResNet}, \textit{SqueezeNet}, \textit{RegNet}, \textit{MnasNet}, \textit{MobileNet}, \textit{EfficientNet}, \etc.) and vision transformers (\eg~\textit{ViT}, \textit{Swin Transformer}, and \textit{MaxViT}), to show that \textbf{I-ASIDE} can not only \textbf{measure} the perturbation robustness but also \textbf{provide interpretations} of its mechanisms. 
\end{chapterabstract}


\lettrine[lines=3]{D}{eep} learning \citep{goodfellow2016deep} has achieved remarkable success across a diversity of domains, including computer vision \citep{krizhevsky2012imagenet,he2016deep,dosovitskiy2020image}, natural language processing \citep{hochreiter1997long,vaswani2017attention,radford2019language,devlin2019bert,brown2020language,openai2023gpt4,deepseekai2025deepseekr1}, and graph representation learning \citep{perozzi2014deepwalk,velivckovic2018graph,xu2018powerful}. These successes have largely expanded the frontiers of what is achievable in artificial intelligence, enabling powerful capabilities across a wide range of tasks, as well as modeling paradigms such as generative modeling \citep{hinton2006reducing,goodfellow2014generative,dinh2017density,ho2020denoising,song2020score,lipman2023flow}. 

Despite these remarkable advances, critical challenges persist, notably in ensuring \textbf{robustness} to small input perturbations and \textbf{generalization} to out-of-distribution data \citep{szegedy2014intriguing,goodfellow2014explaining,hendrycks2019benchmarking,ilyas2019adversarial,recht2019imagenet,luo2024interpreting}. These critical challenges highlight the need to understand the underlying fundamental principles that govern \textbf{robustness} and \textbf{generalization}, and to establish deep learning systems that are \textbf{reliable} --- performing consistently under expected conditions on in-distribution data \citep{nist2023ai}, \textbf{resilient} --- capable of recovering from unexpected conditions such as noise or adversarial attacks \citep{nist2023ai}, and \textbf{trustworthy} --- behaving transparently, ethically, in alignment with intended use, and technically robust \citep{eu-trustworthy-ai-guidelines,eu-ai-act-2024} --- in safety-critical applications.

\begin{figure}[t!]
    \centering
    
    \resizebox{1\linewidth}{!}{
      \begin{tikzpicture}[
  node distance=2cm,
  databox/.style={
    cylinder,
    shape border rotate=90,
    draw,
    minimum width=2cm,
    minimum height=2.5cm
  },
  funcbox/.style={
    rectangle,
    draw,
    thick,
    minimum width=8cm,
    minimum height=4.5cm
  },
  ellipseblock/.style={
    ellipse,
    draw,
    thick,
    minimum width=2.5cm,
    minimum height=2.5cm
  },
  circleblock/.style={
    circle,
    draw,
    thick,
    minimum size=2cm
  },
  arrow/.style={
        -{Classical TikZ Rightarrow[length=3mm]},
        thick
    }
]

  %
  \makeatletter
  \newcommand{\gear}[5]{%
    \foreach \i in {1,...,#1}{%
      [rotate=(\i-1)*360/#1]
        (0:#2) arc (0:#4:#2)
        -- (#4+#5:#3) arc (#4+#5:360/#1-#5:#3)
        -- (360/#1:#2)
    }%
  }
  \makeatother
  %

  \node[databox]                   (data)      {Data: $x \in \mathcal{X}$};
  
  \node[below=0.3cm of data, text width=8cm, align=center,text=Emerald] (lipschitztext) {\textbf{Lipschitz continuity modulates data feature propagation in $f$.}};
  
  \node[funcbox, right=of data] (function) {}; 
  \node at ([xshift=0cm, yshift=1.4cm]function.center) {Learned function: $f(x;\boldsymbol{\theta}): \mathcal{X} \times \Theta \to \tilde{\mathcal{Y}}$};

    \node[
  anchor=north,
  text width=10cm,
  align=center,text=Emerald
    ] at ([yshift=-0.3cm]function.south) 
    {\textbf{Optimization induces the dynamics of Lipschitz continuity in $f$.}};

  \node[ellipseblock, above=of function]     (prior)     {Supervision signal: $y \in \mathcal{Y}$};
  \node[circleblock,  right=of function] (criterion) {Criterion: $\ell_f(f(x;\boldsymbol{\theta}),y)$};

  \draw[arrow] (data)     -- (function);
  \draw[arrow] (function) -- (criterion);
  \draw[arrow] ([yshift=-1cm]data.north)     |- (prior);
  \draw[arrow] (prior)    -| (criterion);

  \begin{scope}[shift={(function.center)}]
    \def\nteeth{18}
    \def\rin{0.8}     
    \def\rout{1}      
    \def\toothang{10} 
    \def\gapang{2}    
    \def\halfPitch{180/\nteeth}

    \coordinate (G1) at (-2*\rout,0);
    \coordinate (G2) at (   0,   -1cm);
    \coordinate (G3) at ( 2*\rout,0);

    \begin{scope}[fill=gray!50, thick]
      \draw[shift={(G1)}] \gear{\nteeth}{\rin}{\rout}{\toothang}{\gapang};
      \draw[shift={(G2)},rotate=\halfPitch]
        \gear{\nteeth}{\rin}{\rout}{\toothang}{\gapang};
      \draw[shift={(G3)}] \gear{\nteeth}{\rin}{\rout}{\toothang}{\gapang};
    \end{scope}

    \node at (G1) {$\boldsymbol{\theta}^{(1)}$};
    \node at (G2) {$\boldsymbol{\theta}^{(2)}$};
    \node at (G3) {$\boldsymbol{\theta}^{(3)}$};
  \end{scope}

\end{tikzpicture}
    }

    \caption[Principles of Lipschitz Continuity Manifest in Paradigm of Machine Learning]{\textbf{Principles of Lipschitz Continuity Manifest in Paradigm of Machine Learning}. In the paradigm of machine learning, we consider \textit{data} $x \in \mathcal{X}$, \textit{supervision signal} $y \in \mathcal{Y}$, a \textit{learned function} $f: \mathcal{X} \times \boldsymbol{\Theta} \to \Tilde{\mathcal{Y}}$ --- where $\Tilde{\mathcal{Y}}$ is output space --- parameterized by $\theta^{(\ell)} \in \mathbb{R}^{d_\ell}$ and $\boldsymbol{\Theta}:=\mathbb{R}^{d_1 \times d_2 \times \cdots \times d_L}$, and a \textit{criterion} $\ell_f: (f(x, \boldsymbol{\theta}), \mathcal{Y}) \to \mathbb{R}$. The Lipschitz constant $K_f$ of $f$ in this paradigm is determined by factors such as the hypothesis class for $f$, learning algorithm, parameters $\boldsymbol{\theta}$, training data $(\mathcal{X}, \mathcal{Y})$, \etc. It also modulates the behavior of $f$ with respect to the input features. This thesis investigates the \textbf{principles of Lipschitz continuity} in this paradigm from a \textbf{complementary dichotomy of perspectives}: an \textbf{internal} perspective --- focusing on the temporal evolution of Lipschitz continuity in neural networks during training; and an \textbf{external} perspective --- examining how Lipschitz continuity modulates the behavior of $f$ with respect to the features in input data, particularly its role in governing frequency signal propagation in neural networks.}

    \label{fig:paradigm_of_machine_learning}

\end{figure}

The paradigm of machine learning \citep{goodfellow2016deep}, illustrated in Figure~\ref{fig:paradigm_of_machine_learning}, typically comprises the \emph{data} $x \in \mathcal{X} \subseteq \mathbb{R}^m$, \emph{supervision signal} $y \in \mathcal{Y} $, a \emph{learned function} $f(x; \boldsymbol{\theta})$ with output space $\Tilde{\mathcal{Y}} \subseteq \mathbb{R}^n$:
\begin{align}
    f: \mathcal{X} \times \boldsymbol{\Theta} \to \Tilde{\mathcal{Y}}
    ,
\end{align}
parameterized by:
\begin{align}
\boldsymbol{\theta}:=\{\boldsymbol{\theta}^{(1)}, \boldsymbol{\theta}^{(2)},\cdots \} \in \boldsymbol{\Theta}
\subseteq \mathbb{R}^p
,
\end{align}
and a \emph{criterion}:
\begin{align}
    \ell_f: (f(x; \boldsymbol{\theta}), y) \to \mathbb{R}.
\end{align}
Note that the \emph{supervision signals} differ across learning paradigms, for example:
\begin{enumerate}[label=(\roman*)]
    \item \textbf{Supervised Learning.} The \emph{supervision signals} are explicitly provided as \emph{labels} by the dataset;

    \item \textbf{Self-Supervised Learning.} The \emph{supervision signals} are derived from inherent structure or priors in the data --- for example, in contrastive learning, an augmented view of an image can serve as the label for the original image \citep{chen2020simple};

    \item \textbf{Reinforcement Learning.} The \emph{supervision signals} are manifested as \emph{reward signals} from the interactions with the environment, which are often noisy, sparse, and delayed \citep[Preface \& \S~1]{sutton1998reinforcement}.

\end{enumerate}

Among the theoretical tools available, \textbf{Lipschitz continuity} \citep[Theorem 9.19, \S~9]{rudin1976principles} plays a pivotal role in governing the fundamental properties of the function $f$ in this paradigm related to robustness and generalization. The \emph{Lipschitz constant} $K_f$ of $f$, with respect to the $\ell_2$-norm, defined as:
\begin{align}
    K_f := \sup_{\forall~u \neq v} \frac{\|f(u) - f(v)\|_2}{\|u - v\|_2},
\end{align}
measures the worst-case sensitivity of the network to small input perturbations. Equivalently, it bounds the maximum rate of change of $f$ with respect to its inputs in the $\ell_2$-norm \citep[Theorem 9.19, \S~9]{rudin1976principles}:
\begin{align}
K_f \geq \sup_x \|\nabla_x f(x) \|_2
    ,
\end{align}
where $\nabla_x (\cdot)$ denotes a differential operator acting on $(\cdot)$ with respect to $x$. A lower Lipschitz constant $K_f$ ensures a smoother mapping $f$, enhancing the network's ability to generalize to unseen data. Moreover, Lipschitz continuity forms a theoretical foundation for \textbf{certifiable robustness} --- where safety guarantees under bounded perturbations can be formally proven --- in safety-critical applications. 

While Lipschitz continuity is crucial in governing the fundamental properties of neural networks related to robustness and generalization, prior research has largely concentrated on empirical aspects, such as regularization techniques \citep{miyato2018spectral,arjovsky2017wasserstein,cisse2017parseval,anil2019sorting}, leaving the underlying principles less explored. This thesis is motivated by the absence of theoretical underpinnings of the principles of Lipschitz continuity in neural networks --- a foundation that is critical for establishing reliable, resilient, and trustworthy deep learning systems. To address this gap, it seeks to advance a principled understanding of the \textbf{principles of Lipschitz continuity in neural networks} within the paradigm of machine learning, examined from \textbf{a complementary dichotomy of perspectives}: 
\begin{enumerate}[label={}]
    \item an \textbf{internal} perspective --- focusing on the temporal evolution of Lipschitz continuity in neural networks during training (\ie, \textbf{Training Dynamics});

    \item an \textbf{external} perspective --- investigating how Lipschitz continuity modulates the behavior of neural networks with respect to features in the input data, particularly its role in governing frequency signal propagation (\ie, \textbf{Modulation of Frequency Signal Propagation}).
\end{enumerate}
Together, these complementary perspectives form the basis of the primary research questions presented in Section~\ref{sec:research_questions} and underscore the principles under investigation within this thesis.

\section{Research questions}
\label{sec:research_questions}


This thesis is centered around an overarching inquiry from a complementary dichotomy of \textbf{internal} and \textbf{external} perspectives:
\begin{quote}
\textit{What are the \textbf{principles of Lipschitz continuity in neural networks}, from its temporal evolution during training (an \textbf{internal perspective}) to its modulation of frequency signal propagation (an \textbf{external perspective})?}
\end{quote}

To support this central inquiry, it investigates three primary research questions (RQs) with a structured approach, each gives rise to a set of secondary research questions (SRQs), which are explored in one or two dedicated chapters.

\begin{questionbox}{(RQ1) State of Knowledge}
\begin{treelist}
    \item (RQ1) \textbf{State of Knowledge} --- what is the state of knowledge of Lipschitz continuity in deep learning? 
    
    \begin{treelist}
        \item (SRQ1) \textbf{Theoretical Foundations} --- what are the theoretical foundations of Lipschitz continuity in deep learning? 
        
        \item (SRQ2) \textbf{Regularization Mechanism} --- how does the constraint of Lipschitz continuity serve as a regularization mechanism for improving robustness and generalization capabilities in deep learning?
        
    \end{treelist}
\end{treelist}
\end{questionbox}

\begin{questionbox}{(RQ2) Training Dynamics --- An Internal Perspective}
\begin{treelist}
    \item (RQ2) \textbf{Training Dynamics} --- how does Lipschitz continuity of neural networks evolve over time during training?
    \begin{treelist}
        \item (SRQ3) \textbf{Parameter Space Perturbation} --- how do the spectral properties (\eg~singular values) of parameter matrices change in training? During training, the optimization process updates parameter matrices, thereby altering the spectral structures accordingly. 

        \item (SRQ4) \textbf{Stochastic Dynamic Analysis of Lipschitz Continuity} --- how does Lipschitz continuity evolve over time during training in neural networks? 
        
    \end{treelist}
\end{treelist}
\end{questionbox}

\begin{questionbox}{(RQ3) Modulation of Frequency Signal Propagation --- An External Perspective}
\begin{treelist}
    \item (RQ3) \textbf{Modulation of Frequency Signal Propagation} --- how does Lipschitz continuity modulate the frequency signal propagation in neural networks, and how does it help us for understanding the global robustness of image models?
    \begin{treelist}
        \item (SRQ5) \textbf{Fourier Analysis of Lipschitz Continuity} --- how does Lipschitz continuity modulate frequency signal propagation in image models?
        
        \item (SRQ6) \textbf{Interpreting Global Robustness} --- how can global robustness be explained through the lens of spectral bias in Lipschitz continuity? 
    \end{treelist}
\end{treelist}
\end{questionbox}

\section{Thesis Presentation Format}

This thesis compiles and presents a selection of research articles --- published or under review as of the date of submission --- into a \underline{\textbf{self-contained}}, \underline{\textbf{modular}}, \underline{\textbf{coherent}}, \underline{\textbf{monograph-style}} presentation format, all centered on the principles of Lipschitz continuity in neural networks. The research was conducted cumulatively during the Ph.D. training. Portions of the chapter content, which address the research questions, are reused either verbatim or with minor modifications from the selected articles, in accordance with institutional guidelines and the policies of the relevant publishers. Each core chapter is presented in the style of a standalone journal article, including an \textbf{abstract}, \textbf{introduction}, and other standard components in journal articles. For consistency with the original publications, the term \textbf{paper} is used throughout to refer to the current chapter, and the literature review may include results from our research, either published or available as preprints. Permissions for reuses have been obtained from all co-authors in all articles.

\section{Content Organization}

\begin{figure}[t!]
    \centering
    
    \resizebox{1\linewidth}{!}{





\begin{tikzpicture}[node distance=10mm and 15mm, >=Stealth,
  rq1/.style={
        rectangle, draw, rounded corners, 
        fill=green!20, 
        align=center, text width=5cm, minimum height=2cm},
  rq2/.style={
        rectangle, draw, rounded corners, 
        fill=orange!20, 
        align=center, text width=5cm, minimum height=2cm},
  rq3/.style={
        rectangle, draw, rounded corners, 
        fill=cyan!20, 
        align=center, text width=5cm, minimum height=2cm},
  legendbox/.style={
        rectangle, draw, rounded corners, minimum width=12mm, minimum height=12mm, inner sep=0pt},
  legend label/.style={
        anchor=west, 
        }
]

  \begin{scope}[local bounding box=diagram]
  
    \node[rq1] (LiteratureReview) {\hyperref[chapter:literature_review]{Chapter~\ref{chapter:literature_review}: Literature Review}};

    \node[rq1, ellipse, draw, align=center, text width=3.5cm, minimum height=2cm, right=of LiteratureReview, yshift=15mm] (TheoreticalFoundations) {Theoretical Foundations};

    \node[rq1, ellipse, draw, align=center, text width=3.5cm, minimum height=2cm, right=of LiteratureReview, yshift=-15mm] (Regularizations) {Regularizations};

    \node[rq2, above=of TheoreticalFoundations] (PerturbationAnalysis) {\hyperref[chapter:perturbation_analysis]{Chapter~\ref{chapter:perturbation_analysis}: Operator-Theoretic Perturbation Analysis}};

    \node[rq2, ellipse, draw, align=center, text width=3.5cm, minimum height=2cm, left=of PerturbationAnalysis, yshift=15mm] (PerturbationTheory) {Perturbation Theory};

    \node[rq2, ellipse, draw, align=center, text width=3.5cm, minimum height=2cm, left=of PerturbationAnalysis, yshift=-15mm] (OperatorTheory) {Operator Theory};

    \node[rq2, right=of TheoreticalFoundations] (InducedDynamics) {\hyperref[chapter:optimization_induced_dynamics]{Chapter~\ref{chapter:optimization_induced_dynamics}: Stochastic Dynamic Analysis of Lipschitz Continuity}};

    \node[rq2, ellipse, draw, align=center, text width=3.5cm, minimum height=2cm, above=of InducedDynamics] (SDEFoundations) {Matrix-Valued SDE};

    \node[rq2, ellipse, draw, align=center, text width=3.5cm, minimum height=2cm, below=of InducedDynamics] (SGDOptimization) {SGD Optimization};

    \node[rq2, ellipse, draw, align=center, text width=3.5cm, minimum height=2cm, right=of InducedDynamics] (TheoreticalImplications) {Theoretical Implications in Training Dynamics};

    \node[rq2, ellipse, draw, align=center, text width=3.5cm, minimum height=2cm, right=of TheoreticalImplications] (NoisyGradientRegularizations) {Noisy Gradient Regularizations};

    \node[rq2, ellipse, draw, align=center, text width=3.5cm, minimum height=2cm, above=of NoisyGradientRegularizations] (NearConvergenceBehaviours) {Near Convergence Behaviours};

    \node[rq2, ellipse, draw, align=center, text width=3.5cm, minimum height=2cm, above=of NearConvergenceBehaviours] (ParameterInitialization) {Robustness due to Parameter Initialization};

    \node[rq2, ellipse, draw, align=center, text width=3.5cm, minimum height=2cm, below=of NoisyGradientRegularizations] (BatchSize) {Batch Size};

    \node[rq2, ellipse, draw, align=center, text width=3.5cm, minimum height=2cm, below=of BatchSize] (SamplingTrajectory) {Mini-Batch Sampling Trajectory};
    
    \node[rq3, below=of Regularizations] (LocalLipschitz) {\hyperref[chapter:fourier_analysis_of_lipschitz_continuity]{Chapter~\ref{chapter:fourier_analysis_of_lipschitz_continuity}: Fourier Analysis of Lipschitz Continuity}};

    \node[rq3, right=of LocalLipschitz] (GlobalInterpretation) {\hyperref[chapter:interpreting_global_robustness]{Chapter~\ref{chapter:interpreting_global_robustness}: Interpreting Global Robustness}};

    \node[rq3, ellipse, draw, align=center, text width=3.5cm, minimum height=2cm, below=of GlobalInterpretation, xshift=-30mm] (GameTheory) {Shapley Value Theory};

    \node[rq3, ellipse, draw, align=center, text width=3.5cm, minimum height=2cm, below=of GlobalInterpretation, xshift=30mm] (InformationTheory) {Information Theory};

    \draw[->] (LiteratureReview) -- (TheoreticalFoundations);
    \draw[->] (LiteratureReview) -- (Regularizations);
    \draw[->] (OperatorTheory) -- (PerturbationAnalysis);
    \draw[->] (PerturbationTheory) -- (PerturbationAnalysis);
    \draw[->] (TheoreticalFoundations) -- (InducedDynamics);
    \draw[->] (SGDOptimization) -- (InducedDynamics);
    \draw[->] (SDEFoundations) -- (InducedDynamics);
    \draw[->] (PerturbationAnalysis) -- (InducedDynamics);
    \draw[->] (InducedDynamics) -- (TheoreticalImplications);
    
    \draw[->] (TheoreticalImplications) -- (NoisyGradientRegularizations);
    \draw[->] (TheoreticalImplications) -- (NearConvergenceBehaviours);
    \draw[->] (TheoreticalImplications) -- (ParameterInitialization);
    \draw[->] (TheoreticalImplications) -- (BatchSize);
    \draw[->] (TheoreticalImplications) -- (SamplingTrajectory);
    
    \draw[->] (Regularizations) -- (LocalLipschitz);
    \draw[->] (LocalLipschitz) -- (GlobalInterpretation);
    \draw[->] (GameTheory) -- (GlobalInterpretation);
    \draw[->] (InformationTheory) -- (GlobalInterpretation);

    \path (current bounding box.north east) coordinate (bbne);
    \path (current bounding box.south west) coordinate (bbsw);
    \path[use as bounding box] (bbsw) rectangle (bbne);
    
  \end{scope}

  \begin{scope}[local bounding box=legend, 
                above=of diagram,
                anchor=south,
                xshift=50mm,
                yshift=90mm
                ]
    \node[legendbox, fill=green!20] (legC) {};
    \node[legend label, right=1mm of legC] (labC) {RQ1: \textbf{State of Knowledge}};
    
    \node[legendbox, fill=orange!20, right=5cm of legC] (legO) {};
    \node[legend label, right=1mm of legO] (labO) {RQ2: \textbf{Training Dynamics}};
    
    \node[legendbox, fill=cyan!20, right=5cm of legO] (legG) {};
    \node[legend label, right=1mm of legG] (labG) {RQ3: \textbf{Modulation of Frequency Signal Propagation}};
  \end{scope}
  
\end{tikzpicture}

    }
    
    \caption[Thesis Thematic Structure]{\textbf{Thesis Thematic Structure}. A high-level conceptual diagram outlining the thematic connections between chapters. This thesis investigates the principles of Lipschitz continuity in neural networks through a structured approach guided by three primary research questions. Each primary question gives rise to two secondary research questions and is addressed in one or two dedicated chapters.}

    \label{fig:thesis_thematic_structure}
\end{figure}


The research questions collectively establish a cohesive thematic investigation into the principles of Lipschitz continuity in neural networks. Figure~\ref{fig:thesis_thematic_structure} illustrates how each question aligns with the core chapters and jointly guides the development of the overarching theoretical framework. The structure of the thesis is outlined below.

\paragraph{(RQ1) State of Knowledge.} What is the state of knowledge of Lipschitz continuity in deep learning? This research question is addressed by surveying the landscape of Lipschitz continuity in deep learning, covering its theoretical foundations, estimation methods, regularization approaches, and role in certifiable robustness. This research question is addressed by surveying the landscape of Lipschitz continuity in deep learning, covering its theoretical foundations, estimation methods, regularization approaches, and role in certifiable robustness. 
\begin{enumerate}[secondary-research-questions,series=SecondaryResearchQuestions,left=2em]
        \item \textbf{Theoretical Foundations} (\hyperref[chapter:literature_review]{Chapter~\ref{chapter:literature_review}: Literature Review}) --- what are the theoretical foundations of Lipschitz continuity in deep learning? This is answered in \hyperref[chapter:literature_review]{Chapter~\ref{chapter:literature_review}: Literature Review}, which is based on an article:
        \begin{enumerate}[label=(\roman*),series=CompiledPublications]
            \item \textcolor{Emerald}{\fullcite{luo2025lipschitzsurvey}}.
        \end{enumerate}
    
        \item \textbf{Regularization Mechanism} (\hyperref[chapter:literature_review]{Chapter~\ref{chapter:literature_review}: Literature Review}) --- how does Lipschitz continuity serve as a regularization mechanism for improving robustness and generalization capabilities in deep learning? This is answered in \hyperref[chapter:literature_review]{Chapter~\ref{chapter:literature_review}: Literature Review}, which is based on an article:
        \begin{enumerate}[label=(\roman*),series=CompiledPublications]
            \item \textcolor{Emerald}{\fullcite{luo2025lipschitzsurvey}}.
        \end{enumerate}
    \end{enumerate}

\paragraph{(RQ2) Training Dynamics.} How does Lipschitz continuity evolve over time during the training in neural networks? To the best of our knowledge, as of the date of submission, this constitutes the first theoretical framework to characterize the temporal evolution of Lipschitz continuity in neural networks leveraging perturbation theory \citep{kato1995perturbation,luo2025spectralvariation} and a system of stochastic differential equations (SDEs). 
\begin{enumerate}[secondary-research-questions,resume=SecondaryResearchQuestions,left=2em]
        \item \textbf{Parameter Space Perturbation} (\hyperref[chapter:perturbation_analysis]{Chapter~\ref{chapter:perturbation_analysis}: Operator-Theoretic Perturbation Analysis}) --- how do the spectral properties (\ie~singular values) of parameter matrices change in training? During training, the optimization process updates parameter matrices, thereby altering the spectral structures accordingly. This is addressed in \hyperref[chapter:perturbation_analysis]{Chapter~\ref{chapter:perturbation_analysis}: Operator-Theoretic Perturbation Analysis}, which is based on an article:
        \begin{enumerate}[resume=CompiledPublications,label=(\roman*)]
            \item \textcolor{Emerald}{\fullcite{luo2025spectralvariation}}.
        \end{enumerate}

        \item \textbf{Stochastic Dynamic Analysis of Lipschitz Continuity} (\hyperref[chapter:optimization_induced_dynamics]{Chapter~\ref{chapter:optimization_induced_dynamics}: Stochastic Dynamic Analysis of Lipschitz Continuity}) --- how does Lipschitz continuity evolve over time during training in neural networks? This is answered in \hyperref[chapter:optimization_induced_dynamics]{Chapter~\ref{chapter:optimization_induced_dynamics}: Stochastic Dynamic Analysis of Lipschitz Continuity}, which is based on an article:
        \begin{enumerate}[resume=CompiledPublications,label=(\roman*)]
            \item \textcolor{Emerald}{\fullcite{luo2025lipschitz}}.
        \end{enumerate}
    
\end{enumerate}

\paragraph{(RQ3) Modulation of Frequency Signal Propagation.} How does Lipschitz continuity modulate the frequency signal propagation in neural networks, and how does it help us for understanding the global robustness of image models? 
\begin{enumerate}[secondary-research-questions,resume=SecondaryResearchQuestions,left=2em]
        \item \textbf{Fourier Analysis of Lipschitz Continuity} (Chapter~\ref{chapter:fourier_analysis_of_lipschitz_continuity}: Fourier Analysis of Lipschitz Continuity) --- how does Lipschitz continuity modulate frequency signal propagation in image models? This is answered in \hyperref[chapter:fourier_analysis_of_lipschitz_continuity]{Chapter~\ref{chapter:fourier_analysis_of_lipschitz_continuity}: Fourier Analysis of Lipschitz Continuity}. 
        
        \item \textbf{Interpreting Global Robustness} (\hyperref[chapter:interpreting_global_robustness]{Chapter~\ref{chapter:interpreting_global_robustness}: Interpreting Global Robustness}) --- how can global robustness be explained through the lens of spectral bias in Lipschitz continuity? This is answered in \hyperref[chapter:interpreting_global_robustness]{Chapter~\ref{chapter:interpreting_global_robustness}: Interpreting Global Robustness}, which is based on an article:
        \begin{enumerate}[resume=CompiledPublications,label=(\roman*)]
            \item \textcolor{Emerald}{\fullcite{luo2024interpreting}}.
        \end{enumerate}
\end{enumerate}

\section{Contributions}

The main contributions are summarized as follows:

\begin{enumerate}[label=(\roman*)]
    \item \textbf{Comprehensive literature study.} We present an extensive survey on Lipschitz continuity in deep learning, covering theoretical foundations, estimation methods, regularization approaches, and certifiable robustness. This survey serves to consolidate existing knowledge.

    \item \textbf{Operator-Theoretic Perturbation Analysis.} We develop a formal operator-theoretic framework for analyzing spectral variations of parameter matrices in training. This framework enables precise characterization of parameter matrices during training.

    \item \textbf{Stochastic analysis of Lipschitz dynamics.} We introduce a theoretical framework that characterizes the temporal evolution of Lipschitz continuity of neural networks during training. This framework is formulated using a system of stochastic differential equations (SDEs) and enables a rigorous analysis of how Lipschitz continuity evolves throughout training. Our analysis identifies key factors that govern the temporal evolution of Lipschitz continuity of neural networks.

    \item \textbf{Interpreting Global Robustness.} We begin by analyzing Lipschitz continuity through the lens of Fourier analysis, revealing how it modulates frequency signal propagation in neural networks. Building on this foundation, we introduce a novel perspective for interpreting the global robustness of image models --- incorporating information theory and Shapley value theory --- by linking the spectral bias of Lipschitz continuity to robustness against input perturbations.
    
\end{enumerate}

\newpage
\begin{subappendices}
\section{Appendix}

\section{Fairness division axioms}
\label{app:decomp_axioms}

\textbf{\textit{Symmetry} axiom}: Let $\widetilde{\mathcal{I}} \in 2^\mathcal{I}$ be some spectral player coalition. For $\forall~\mathcal{I}_i,\mathcal{I}_j \in \mathcal{I} \land \mathcal{I}_i,\mathcal{I}_j \notin \widetilde{\mathcal{I}}$, the statement $v(\widetilde{\mathcal{I}} \cup \{\mathcal{I}_i\}) = v(\widetilde{\mathcal{I}} \cup \{\mathcal{I}_j\})$ implies $\psi_i(\mathcal{I},v) = \psi_j(\mathcal{I},v)$. This axiom restates the statement `\textit{equal treatment of equals}' principle mathematically. This axiom states that the `names' of players should have no effect on the `treatments' by the characteristic function in coalition games \citep{roth1988shapley}.

\textbf{\textit{Linearity} axiom}: Let $u$ and $v$ be two characteristic functions. Let $(\mathcal{I}, u)$ and $(\mathcal{I}, v)$ be two coalition games. Let $(u + v)(\widetilde{\mathcal{I}}) := u(\widetilde{\mathcal{I}}) + v(\widetilde{\mathcal{I}})$ where $\widetilde{\mathcal{I}} \in 2^\mathcal{I}$. The divisions of the new coalition game $(\mathcal{I}, u + v)$ should satisfy: $\psi_i(\mathcal{I}, u + v) = \psi_i(\mathcal{I}, u) + \psi_i(\mathcal{I}, v)$. This axiom is also known as `\textit{additivity} axiom' and guarantees the uniqueness of the solution of dividing payoffs among players \citep{roth1988shapley}.

\textbf{\textit{Efficiency} axiom}: This axiom states that the sum of the divisions of all players must be summed to the worth of the player set (the grand coalition): 
$\sum\limits_{i=0}^{M-1} \psi_i(\mathcal{I},v) = v(\mathcal{I})$.

\textbf{\textit{Dummy player} axiom}: A dummy player (null player) $\mathcal{I}_*$ is the player who has no contribution such that: $\psi_*(\mathcal{I},v)=0$ and $v(\widetilde{\mathcal{I}} \cup \{\mathcal{I}_*\}) \equiv v(\widetilde{\mathcal{I}})$ for $\forall~\mathcal{I}_* \notin \widetilde{\mathcal{I}} \land \mathcal{I}_* \subseteq \mathcal{I}$.

\begin{remark}
In the literature \citep{roth1988shapley}, the \textit{efficiency} axiom and the \textit{dummy player} axiom are also combined and relabeled as \textit{carrier} axiom.     
\end{remark}

\newpage
\subsection{Spectral signal-to-noise ratio (SNR)}
\label{app:image_ssnr}

\textbf{Discrete Fourier Transform}. The notion `frequency' measures how `fast' the outputs can change with respect to inputs. High frequency implies that small variations in inputs can cause large changes in outputs. In terms of images, the `inputs' are the pixel spatial locations while the `outputs' are the pixel values.

Let $x: (i, j) \mapsto \mathbb{R}$ be some 2D image with dimension $M \times N$ which sends every location $(i, j)$ to some real pixel value where $(i, j) \in [M] \times [N]$. Let $\mathscr{F}: \mathbb{R}^2 \mapsto \mathbb{C}^2$ be some DFT functional operator. The DFT of $x$ is given by:
\begin{align}
    \label{equ:dft_def}
    \mathscr{F}(x)(u,v) = 
    \sum_{j=0}^{N-1} \sum_{i=0}^{M-1} x(i, j) e^{- \mathbb{i} 2\pi (\frac{u}{M} i + \frac{v}{N} j) } .
\end{align}

\textbf{Point-wise energy spectral density (ESD)}. 
The ESD measures the energy quantity at a frequency. To simplify discussions, we use \textit{radial frequency}, which is defined as the radius $r$ with respect to zero frequency point (\ie~the frequency center). The energy is defined as the square of the frequency magnitude according to Parseval's Power Theorem.

Let $L_r$ be a circle with radius $r$ on the spectrum of image $x$, as illustrated in Figure~\ref{fig:robust_features_vs_non_robust_features}. The $r$ is referred to as radial frequency. The point-wise ESD function is given by:
\begin{align}
    ESD_r(x):= \frac{1}{|L_r|} \cdot \sum\limits_{(u,v) \in L_r} |\mathscr{F}(x)(u,v)|^2
\end{align}
where $(u,v)$ is the spatial frequency point and $|L_r|$ is the circumference of $L_r$.

\textbf{Spectral signal-to-noise ratio (SNR)}. The SNR can quantify signal robustness. We define the spectral SNR at radius frequency $r$ as:
\begin{align}
    SNR(r) := 
    \frac{ESD_r(x)}{ESD_r(\Delta x)}
\end{align}
where $\Delta x$ is some perturbation. We have characterized the SNRs of some corroptions and adversarial attacks in Figure~\ref{fig:snr_char}.

\newpage
\subsection{Absence assignment scheme}
\label{app:absence_assignment}

There exist multiple choices for the assignments of the absences of spectral layers in coalition filtering design: (1) Assigning to constant zeros (Zeroing), (2) assigning to complex Gaussian noise (Complex Gaussian) and (3) assigning to the corresponding frequency components randomly sampled from other images at the same dataset (Replacement). 

\textbf{Zeroing}. The $\vb*b$ in \eqref{equ:feature_filtering_operator} is set to zeros. 

\textbf{Complex Gaussian}. The $\vb*b$ in \eqref{equ:feature_filtering_operator} is sampled from a \textit{i.i.d.} complex Gaussian distribution: $\mathcal{N}(\mu, \frac{\sigma^2}{2}) + i \mathcal{N}(\mu, \frac{\sigma^2}{2})$. 

\textbf{Replacement}. The $\vb*b$ in \eqref{equ:feature_filtering_operator} is set to: $\vb*b = \mathscr{F}(x^*)$ (where $x^* \thicksim \mathcal{X}$ is a randomly sampled image from some set $\mathcal{X}$). 

\begin{figure}[H]
 

  \centering

\begin{minipage}[t]{0.15\textwidth}
     \centering 
     \subfloat[Zeroing]{
      \includegraphics[width=1\textwidth,height=1\textwidth]{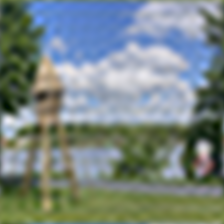}
     }
  \end{minipage}
  \hspace*{\fill}%
     \begin{minipage}[t]{0.15\textwidth}
     \centering 
     \subfloat[Complex Gaussian]{
      \includegraphics[width=1\textwidth,height=1\textwidth]{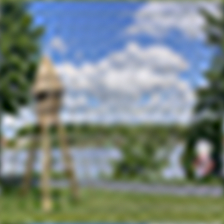}
     }
  \end{minipage}
   \hspace*{\fill}%
     \begin{minipage}[t]{0.15\textwidth}
     \centering 
     \subfloat[Replacement]{
      \includegraphics[width=1\textwidth,height=1\textwidth]{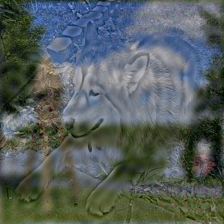}
     }
  \end{minipage}
  \hspace*{\fill}%
     \begin{minipage}[t]{0.15\textwidth}
     \centering 
     \subfloat[Zeroing]{
      \includegraphics[width=1\textwidth,height=1\textwidth]{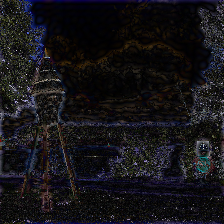}
     }
  \end{minipage}
    \hspace*{\fill}%
     \begin{minipage}[t]{0.15\textwidth}
     \centering 
     \subfloat[Complex Gaussian]{
      \includegraphics[width=1\textwidth,height=1\textwidth]{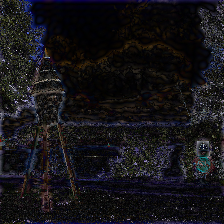}
     }
  \end{minipage}
  \hspace*{\fill}%
     \begin{minipage}[t]{0.15\textwidth}
     \centering 
     \subfloat[Replacement]{
      \includegraphics[width=1\textwidth,height=1\textwidth]{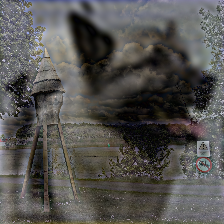}
     }
  \end{minipage}

  \begin{minipage}[t]{0.32\textwidth}
     \centering 
     \subfloat[{\textit{resnet18} w/ Zeroing}]{
      \includegraphics[width=1\textwidth,height=0.5\textwidth]{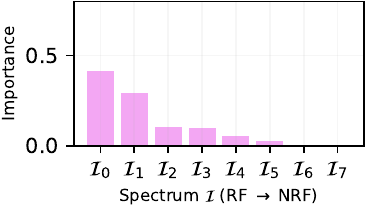}
     }
  \end{minipage}
\hspace*{\fill}%
     \begin{minipage}[t]{0.32\textwidth}
     \centering 
     \subfloat[{\textit{resnet18} w/ Complex Gaussian}]{
      \includegraphics[width=1\textwidth,height=0.5\textwidth]{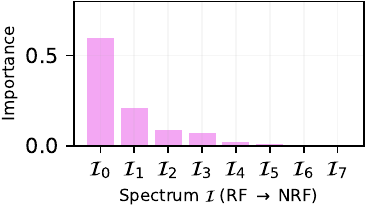}
     }
  \end{minipage}
\hspace*{\fill}%
     \begin{minipage}[t]{0.32\textwidth}
     \centering 
     \subfloat[{\textit{resnet18} w/ Replacement}]{
      \includegraphics[width=1\textwidth,height=0.5\textwidth]{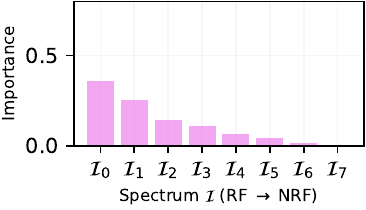}
     }
  \end{minipage}


    \begin{minipage}[t]{0.32\textwidth}
     \centering 
     \subfloat[{\textit{efficientnet\_v2\_s} w/ Zeroing}]{
      \includegraphics[width=1\textwidth,height=0.5\textwidth]{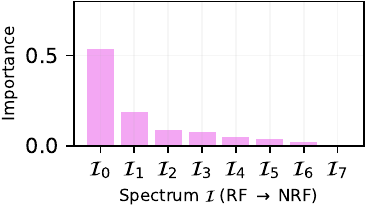}
     }
  \end{minipage}
\hspace*{\fill}%
     \begin{minipage}[t]{0.32\textwidth}
     \centering 
     \subfloat[{\textit{efficientnet\_v2\_s} w/ Complex Gaussian}]{
      \includegraphics[width=1\textwidth,height=0.5\textwidth]{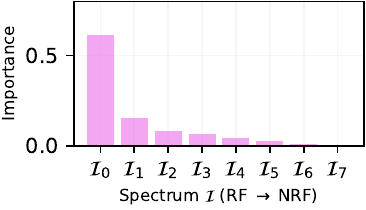}
     }
  \end{minipage}
    \hspace*{\fill}%
     \begin{minipage}[t]{0.32\textwidth}
     \centering 
     \subfloat[{\textit{efficientnet\_v2\_s} w/ Replacement}]{
      \includegraphics[width=1\textwidth,height=0.5\textwidth]{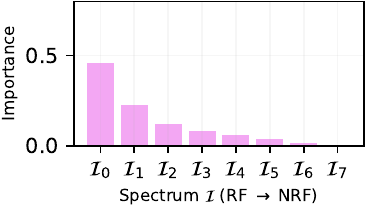}
     }
  \end{minipage}

 
  \caption[Three Absence Assignment Strategies in Spectral Game]{Three absence assignment strategies: (1) Assigning the spectral absences with constant zeros (Zeroing), (2) assigning the spevtral absences with Gaussian noise (Complex Gaussian) and (3) randomly sampling spectral components from the same image datasets (Replacement). The standard complex Gaussian distribution is given by: $\mathcal{N}(0,\frac{1}{2}) + i  \mathcal{N}(0, \frac{1}{2})$. The figures (a), (b) and (c) show the coalition filtering results with the spectral coalition: $\{\mathcal{I}_0\}$. The figures (d), (e) and (f) show the coalition filtering results with the spectral coalition: $\{\mathcal{I}_1,\mathcal{I}_2,\mathcal{I}_3,\mathcal{I}_4,\mathcal{I}_5,\mathcal{I}_6,\mathcal{I}_7\}$. The figures (g) to (l) show the examples of the measured spectral importance distributions of a \textit{resnet18} and a \textit{efficientnet\_v2\_s} (both are pre-trained on \textit{ImageNet}) with the three assignment strategies.}
  
  \vspace{-1.5em} 
 
\label{fig:maskout_examples}
\end{figure}

In our implementation, we simply choose `zeroing': $\vb*b = \vb*0$. Figure~\ref{fig:maskout_examples} shows the filtered image examples by using the above three strategies and also show the examples of measured spectral importance distributions. Empirically, the three strategies have rather similar performance. In this research, we do not unfold the discussions regarding the masking strategy choices. 

\newpage
\subsection{Proof for Spectral Coalition Information Identity Theorem}
\label{app:theo_char_func}

\begin{proof}[Proof for Spectral Coalition Information Identity]

Suppose the probability measures $P(x)$, $P(x,y)$, $P(y|x)$, and $Q(y|x)$ are absolutely continuous with respect to $x$ on domain $\mathcal{X} \bowtie \widetilde{\mathcal{I}}$.

\begin{align}
    \mathbb{I}(\mathcal{X} \bowtie \widetilde{\mathcal{I}}, \mathcal{Y}) &= \int\limits_{\mathcal{X} \bowtie \widetilde{\mathcal{I}}} \sum_{y \in \mathcal{Y}} P(x,y) \cdot \log \frac{P(x, y)}{P(x) \cdot P(y)} dx \\
    &= \int\limits_{\mathcal{X} \bowtie \widetilde{\mathcal{I}}} \sum_{y \in \mathcal{Y}} P(x,y) \cdot \log \left( \frac{P(y|x)\cdot P(x)}{P(y) \cdot P(x)} \cdot \frac{Q(y|x)}{Q(y|x)}\right) dx \\
    &= \int\limits_{\mathcal{X} \bowtie \widetilde{\mathcal{I}}} \sum_{y \in \mathcal{Y}} P(x,y) \cdot \log \left( \frac{P(y|x)}{Q(y|x)} \cdot \frac{1}{P(y)} \cdot Q(y|x) \right) dx  \\
    &= \int\limits_{\mathcal{X} \bowtie \widetilde{\mathcal{I}}} P(x) \left( \sum_{y \in \mathcal{Y}} P(y|x) \cdot \log \frac{P(y|x)}{Q(y|x)}\right) dx  \\
    &\quad\quad -\sum_{y \in \mathcal{Y}} \left( \int\limits_{\mathcal{X} \bowtie \widetilde{\mathcal{I}}} P(x,y) dx \right) \log P(y)\\
    &\quad\quad + \int\limits_{\mathcal{X} \bowtie \widetilde{\mathcal{I}}}  \sum_{y \in \mathcal{Y}} P(x, y) \cdot \log Q(y|x) dx  \\
    &= \mathop\mathbb{E}\limits_{x\in \mathcal{X} \bowtie \widetilde{\mathcal{I}}} \underbrace{KL(P(y|x) || Q(y||x))}_{\mathrm{point-wise}} + H(\mathcal{Y}) + \int\limits_{\mathcal{X} \bowtie \widetilde{\mathcal{I}}}  P(x) \left(\sum_{y \in \mathcal{Y}} P(y|x) \cdot \log Q(y|x) \right) dx   \\
    &= \mathop\mathbb{E}\limits_{x\in \mathcal{X} \bowtie \widetilde{\mathcal{I}}} \underbrace{KL(P(y|x) || Q(y||x))}_{\mathrm{point-wise}} + H(\mathcal{Y}) + \mathop\mathbb{E}\limits_{x \in \mathcal{X} \bowtie \widetilde{\mathcal{I}}} \sum_{y \in \mathcal{Y}} P(y|x) \cdot \log Q(y|x) \\
    &=\mathop\mathbb{E}\limits_{x\in \mathcal{X} \bowtie \widetilde{\mathcal{I}}} \underbrace{KL(P(y|x) || Q(y||x))}_{\mathrm{point-wise}} + H(\mathcal{Y}) + v(\widetilde{\mathcal{I}}) + C
\end{align}
where $H(\mathcal{Y})$ is the Shannon entropy of the label set $\mathcal{Y}$.

\end{proof}

\newpage
\section{Information quantity relationship in spectral coalitions}
\label{app:info_quantity_relation}

\begin{figure}[H]
 
      
  \centering

    \includegraphics[width=0.6\columnwidth,]{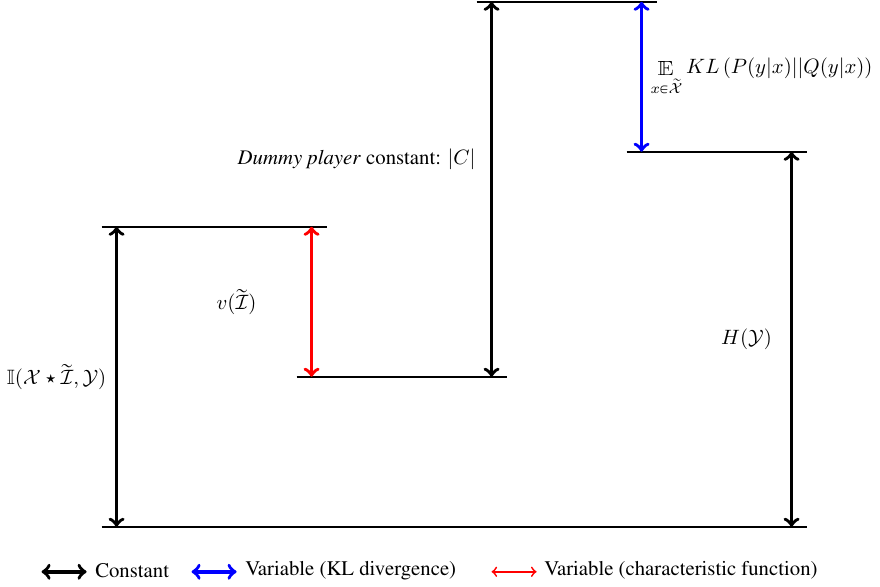}

  \caption[Information Quantity Relationship in Spectral Game]{Information quantity relationship. This shows the theoretical information quantity relationship between what the characteristic function $v$ measures and the mutual information $\mathbb{I}(\mathcal{X} \bowtie \widetilde{\mathcal{I}}, \mathcal{Y})$. For a given coalition $\widetilde{\mathcal{I}}$, a dataset $\langle \mathcal{X}, \mathcal{Y}\rangle$ and a classifier $Q$, the $v$ measures how much information the classifier $Q$ utilizes in decisions. The measured results are then used to compute the marginal contributions of features.}

  
  
\label{fig:char_func}
\end{figure}

\newpage
\subsection{Partitioning spectrum with $\ell_{\infty}$ ball over $\ell_{2}$ ball}
\label{app:linfty_vs_l2_ball}

\begin{figure}[H]
 

  \centering

   \includegraphics[width=1\textwidth,]{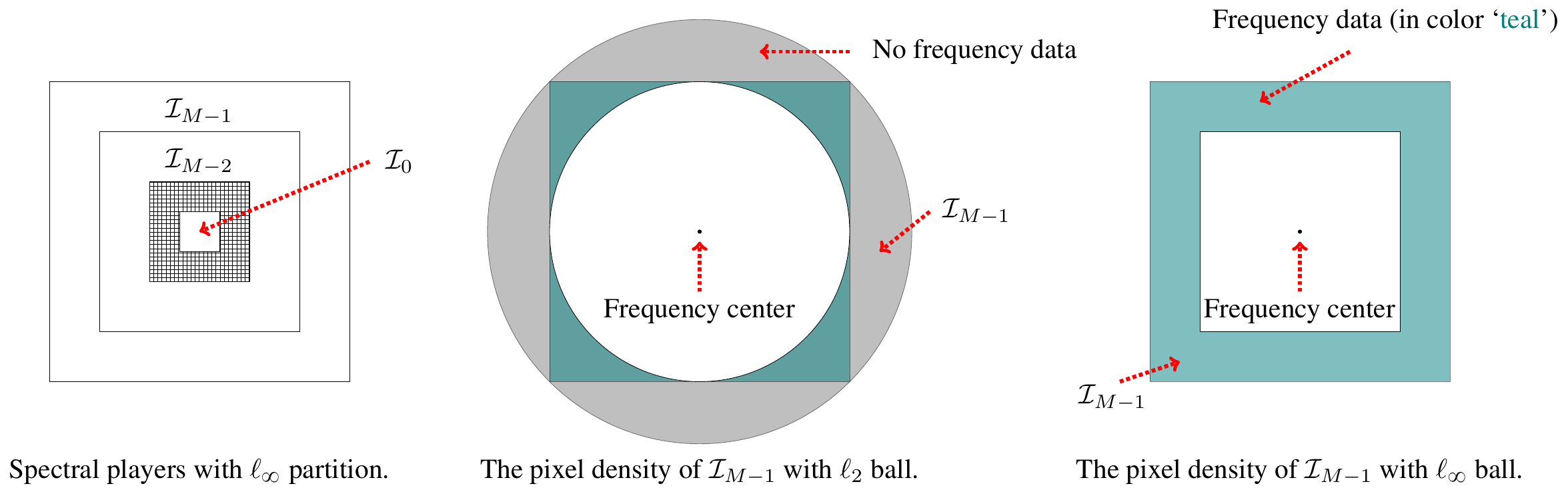}

  \caption[Two Spectral Band Partitioning Schemes in Spectral Game]{Two spectral band partitioning schemes. This shows the motivation we choose $\ell_{\infty}$ ball over $\ell_{2}$ ball in partitioning the frequency domain into the $M$ bands (i.e., $M$ `spectral players') over 2D Fourier spectrum. The frequency data density of the spectral players with $\ell_{\infty}$ remains a constant. However, the frequency data density of the spectral players with $\ell_{2}$ is not a constant since some frequency components do not present. This motives us to empirically choose $\ell_{\infty}$ metric to form spectral players in implementation.}


\label{fig:linfty_vs_l2_ball}
\end{figure}

\newpage
\subsection{Normalizing summarized SIDs}
\label{app:proof_score_formula}

We normalize the above result and set:
\begin{align}
    S(v) &:= \frac{\abs{{\vb*\beta}^T \bar\Psi(v) - \frac{||\vb*\beta||_1}{M}}}{\sup \abs{{\vb*\beta}^T \bar\Psi(v) - \frac{||\vb*\beta||_1}{M}}} \\
    &= \frac{\abs{{\vb*\beta}^T \bar\Psi(v) - \frac{||\vb*\beta||_1}{M}}}{\sup \abs{||\vb*\beta||_2 \cdot || \bar\Psi(v)||_2 - \frac{||\vb*\beta||_1}{M}}} \\
    &= \frac{\abs{{\vb*\beta}^T \bar\Psi(v) - \frac{||\vb*\beta||_1}{M}}}{\abs{||\vb*\beta||_2  - \frac{||\vb*\beta||_1}{M}}} \\
    &= \abs{\frac{{{\vb*{\bar\beta}}}^T \bar\Psi(v) - \frac{1}{M} \frac{||\vb*\beta||_1}{||\vb*\beta||_2}}{1 - \frac{1}{M} \frac{||\vb*\beta||_1}{||\vb*\beta||_2}}}. 
\end{align}
where ${\vb*{\bar\beta}} = \frac{\vb*{\beta}}{||\vb*{\beta}||_2}$ and $\sup \abs{{\vb*\beta}^T \bar\Psi(v) - \frac{||\vb*\beta||_1}{M}}$ is derived by:
\begin{align}
    \sup \abs{{\vb*\beta}^T \bar\Psi(v) - \frac{||\vb*\beta||_1}{M}} &= \abs{\sup {\vb*\beta}^T \bar\Psi(v) - \frac{||\vb*\beta||_1}{M}} \\
    &= \abs{\sup ||{\vb*\beta}||_2 \cdot ||\bar\Psi(v)||_2 - \frac{||\vb*\beta||_1}{M}}~~~~\mathrm{s.t.}~||\bar\Psi(v)||_1 = 1 \\
    &=\abs{||{\vb*\beta}||_2  - \frac{||\vb*\beta||_1}{M}}~~~~\mathrm{since}~||\bar\Psi(v)||_2^2 \leq ||\bar\Psi(v)||_1^2.
\end{align}

Set $\eta=\frac{1}{M}\frac{||\vb*\beta||_1}{||\vb*\beta||_2}$:
\begin{align}
    S(v) = \abs{\frac{{{\vb*{\bar\beta}}}^T \bar\Psi(v) - \eta}{1 - \eta}}.
\end{align}
\qed

\newpage


\subsection{How many samples are sufficient?}
\label{app:error_bound_analaysis}

\begin{figure}[H]
 

  \centering

 \begin{minipage}[t]{0.32\textwidth}
     \centering 
     \subfloat[{\textit{resnet18 @CIFAR10}}]{
      \includegraphics[trim=0in 0in 0in 0in, width=1\textwidth, clip]{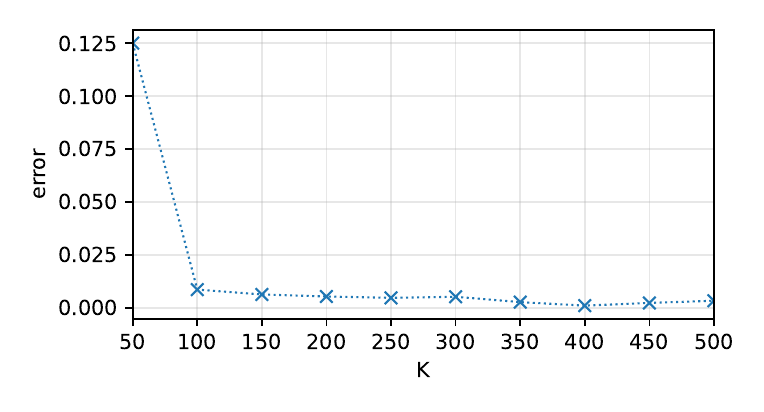}
     }
  \end{minipage}
  \hspace*{\fill}%
   \begin{minipage}[t]{0.32\textwidth}
     \centering 
     \subfloat[{\textit{resnet18 @CIFAR100}}]{
      \includegraphics[trim=0in 0in 0in 0in, width=1\textwidth, clip]{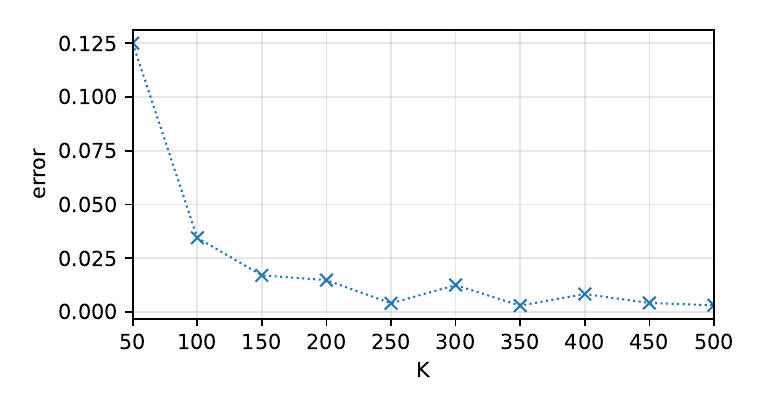}
     }
  \end{minipage}
  \hspace*{\fill}%
   \begin{minipage}[t]{0.32\textwidth}
     \centering 
     \subfloat[{\textit{resnet18 @ImageNet}}]{
      \includegraphics[trim=0in 0in 0in 0in, width=1\textwidth, clip]{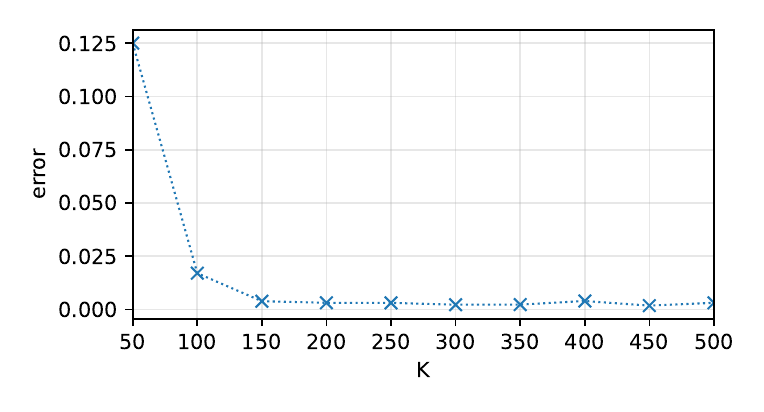}
     }
  \end{minipage}

  
  \caption[Convergence of Relative Estimation Errors in Spectral Game]{Convergence of relative estimation errors converge with respect to the numbers of samples $K$. The errors are measured by: $\frac{1}{M}||\Psi^{(i+1)}(v) - \Psi^{(i)}(v)||_1$ where $\Psi^{(i)}(v)$ denotes the $i$-th measured spectral importance distribution with respect to characteristic function $v$. The experiments are conducted on \textit{CIFAR10}, \textit{CIFAR100} and \textit{ImageNet} with \textit{resnet18}.}


\label{fig:error_estimate}
\end{figure}

\textbf{Error bound analysis}. Let $K$ be the number of the samples of some baseline dataset. Let:
\begin{align}
\Delta v(\Tilde{\mathcal{I}}, \mathcal{I}_i) := v(\Tilde{\mathcal{I}} \cup \{\mathcal{I}_i\}) - v(\Tilde{\mathcal{I}})    
\end{align}
and
\begin{align}
    \Delta v(\mathcal{I}_i) := \left( \Delta v(\Tilde{\mathcal{I}}, \mathcal{I}_i)\right)_{\Tilde{\mathcal{I}} \subseteq \mathcal{I}}
\end{align}
and
\begin{align}
    W := \left( 
    \frac{1}{M}\binom{M - 1}{|\Tilde{\mathcal{I}}|}^{-1}
    \right)_{\Tilde{\mathcal{I}} \subseteq \mathcal{I}}.
\end{align}
Hence:
\begin{align}
    \psi_i(\mathcal{I},v) = W^T \Delta v(\mathcal{I}_i)
\end{align}
where $||W||_1 \equiv 1$ since $W$ is a probability distribution. Let $\bar\psi_i$, $\Delta \bar v(\mathcal{I}_i)$ and $\Delta \bar v(\Tilde{\mathcal{I}}, \mathcal{I}_i)$ be estimations with $K$ samples using Monte Carlo sampling. The error bound with $\ell_1$ norm is given by:
\begin{align}
    \epsilon &\stackrel{\text{def}}{=} \sup_{i} ||\bar\psi_i(\mathcal{I},v) - \psi_i(\mathcal{I},v)||_1  = \sup_{i}  ||W^T\Delta \bar v(\mathcal{I}_i) - W^T\Delta v(\mathcal{I}_i)||_1 \\
    &\leq  \sup_{i} ||W||_1 \cdot || \Delta \bar v(\mathcal{I}_i) - \Delta v(\mathcal{I}_i)||_{\infty} \quad\quad\quad \left(\mathrm{H\"older's~inequality}\right) \\
    &= \sup_{i} || \sum\limits_{\Tilde{\mathcal{I}} \subseteq \mathcal{I} \setminus \mathcal{I}_i} \left( \Delta \bar v(\Tilde{\mathcal{I}}, \mathcal{I}_i) - \Delta v(\Tilde{\mathcal{I}}, \mathcal{I}_i)\right)||_{\infty} \\
    &\leq \sup_{i}  2^{M-1} \cdot \sup_{\Tilde{\mathcal{I}}} || \Delta \bar v(\Tilde{\mathcal{I}}, \mathcal{I}_i) - \Delta v(\Tilde{\mathcal{I}}, \mathcal{I}_i) ||_{\infty} \\
    &=  \sup_{i} 2^{M-1} \cdot \sup_{\Tilde{\mathcal{I}}} || \Delta \bar v(\Tilde{\mathcal{I}}, \mathcal{I}_i) - \Delta v(\Tilde{\mathcal{I}}, \mathcal{I}_i) ||_1 \\
    &\leq  2^{M-1} \cdot \left\{\frac{Var(\Delta \bar v)}{K}\right\}^{\frac{1}{2}}
\end{align}
where $Var(\Delta \bar v)$ gives the upper bound of the variance of $\Delta \bar v(\Tilde{\mathcal{I}}, \mathcal{I}_i)$.

\end{subappendices}